\documentclass[10pt,journal,compsoc]{IEEEtran}

\usepackage{amsmath}
\usepackage{amssymb}
\usepackage{stmaryrd}
\usepackage{booktabs}
\usepackage{url}
\usepackage{algorithm}
\usepackage{algpseudocode}

\usepackage{multicol}
\usepackage{graphicx}
\usepackage[table]{xcolor}
\usepackage{multirow}
\usepackage{diagbox}
\usepackage{subfigure}
\usepackage{float}
\usepackage{color}
\usepackage{framed}

\newcommand{\Mod}[1]{\ (\mathrm{mod}\ #1)}

\DeclareMathOperator*{\argmax}{arg\,max}

\begin{document}

\title{SCEI: A Smart-Contract Driven Edge Intelligence Framework for IoT Systems}

\author{Chenhao~Xu,
        Jiaqi~Ge,
        Yong~Li,
        Yao~Deng,
        Longxiang~Gao,
        Mengshi~Zhang,
        Yong~Xiang,
        and~Xi~Zheng
\thanks{Longxiang~Gao and Xi~Zheng are corresponding authors.}%
\thanks{Chenhao~Xu is with School of Information Technology, Deakin University, Geelong, Australia. E-mail: chenhao.xu@deakin.edu.au.}%
\thanks{Yong~Xiang is with the Deakin Blockchain Innovation Lab, School of Information Technology, Deakin University, Geelong, Australia. E-mail: yong.xiang@deakin.edu.au.}%
\thanks{Jiaqi~Ge is with Jilin University, China. E-mail: gejq18@mails.jlu.edu.cn.}%
\thanks{Yong~Li is with School of Information Technology, Changchun University of Technology, Jilin, China. E-mail: liyong@ccut.edu.cn.}%
\thanks{Yao~Deng and Xi~Zheng are with Macquarie University, NSW, Australia. E-mail: yao.deng@hdr.mq.edu.au and james.zheng@mq.edu.au.}%
\thanks{Longxiang Gao is with Qilu University of Technology and Shandong Computer Science Center, China. E-mail: gaolx@sdas.org.}%
\thanks{Mengshi~Zhang is with Facebook, USA. E-mail: mengshi.zhang@utexas.edu.}%
}

\markboth{IEEE Transactions on Mobile Computing,~Vol.~xx, No.~x, June~2023}%
{Xu \MakeLowercase{\textit{et al.}}: SCEI: A Smart-Contract Driven Edge Intelligence Framework for IoT Systems}

\IEEEtitleabstractindextext{%
\begin{abstract}
Federated learning (FL) enables collaborative training of a shared model on edge devices while maintaining data privacy. FL is effective when dealing with independent and identically distributed (iid) datasets, but struggles with non-iid datasets. Various personalized approaches have been proposed, but such approaches fail to handle underlying shifts in data distribution, such as data distribution skew commonly observed in real-world scenarios (e.g., driver behavior in smart transportation systems changing across time and location). Additionally, trust concerns among unacquainted devices and security concerns with the centralized aggregator pose additional challenges. To address these challenges, this paper presents a dynamically optimized personal deep learning scheme based on blockchain and federated learning. Specifically, the innovative smart contract implemented in the blockchain allows distributed edge devices to reach a consensus on the optimal weights of personalized models. Experimental evaluations using multiple models and real-world datasets demonstrate that the proposed scheme achieves higher accuracy and faster convergence compared to traditional federated and personalized learning approaches.
\end{abstract}

\begin{IEEEkeywords}
Federated Learning, Smart Contract, Personalized Model, Blockchain, Distributed Learning, Deep Learning, IoT.
\end{IEEEkeywords}}

\maketitle

\section{Introduction}

Companies are often hesitant to share their business data with others due to commercial competition and privacy concerns. As a result, an approach that enables collaborators to collectively train a shared model while retaining complete control over their local data becomes crucial to meet various real-world demands. For example, in smart hospital systems, accurate recognition of medical test results, such as blood tests, X-ray tests, and ultrasound tests, heavily relies on image-based classifiers~\cite{srivastava2019deep}. However, the limited amount of local data collected from patients and cases at an individual clinic is inadequate for training a deep-learning model with the desired level of accuracy. Besides, privacy concerns hinder the sharing of disease data across multiple hospitals. Another example is a smart farming system that employs a vision-based plant monitoring system for plant growth management~\cite{boursianis2020internet}. However, privacy concerns pose a challenge as farmers are unwilling to share plant growth data with competitors, resulting in a relatively small amount of local data available for training models on a single farm.

Federated learning (FL) improves model accuracy while preserving privacy by aggregating local gradient updates from training nodes without sharing the local data~\cite{mcmahan2017communication, yang2019federated}. However, the global model trained by FL performs well primarily on local data that follows an independent and identically distributed (iid) pattern, which is rarely encountered in real-world scenarios. The presence of non-iid data, such as diverse patients and cases across different hospitals, poses a challenge to traditional FL: Optimizing the local model on one node will cause an accuracy decline on other nodes~\cite{wang2020optimizing} due to inconsistent data distributions among the local training data. Furthermore, the centralized aggregator in traditional FL is vulnerable to attacks. For instance, even if the local models are encrypted, an attacker who gains access to the centralized aggregation server can exploit information embedded in gradients to reconstruct the original training data through Deep Leakage from Gradients (DLG) attacks~\cite{zhu2019deep}. In addition, the edge devices in FL are unable to trust each other due to commercial competition.

Some researchers proposed schemes aiming to train personalized models in FL~\cite{deng2020adaptive, wu2020personalized}, where the objective is to train models that focus on a subset of the global data distribution~\cite{lee2021opportunistic}. However, such personalized models tend to overfit local data and have difficulty handling data from other nodes that exhibit variances in data distribution (i.e. the data skew problem). For example, a hospital trains a personalized model that accurately predicts the condition of local patients. However, when faced with patients transferred from another hospital, the performance of the personalized model significantly deteriorates. As a result, an orchestrator is required to maintain a balance between the personalized models and the global model for each node to deal with data skew issues.

To improve the security and the credibility of FL, several schemes combining blockchain with FL have been proposed~\cite{xu2021asynchronous, xu2022efficient, warnat2021swarm}. Blockchain is regarded as an effective improvement to FL due to its features such as privacy protection, tamper resistance, and decentralized nature~\cite{xu2021light}. One particular part of blockchain, the \emph{Smart Contract}, is a self-executing decentralized computer program that has often been underutilized. In many cases, it is treated merely as an access interface for distributed storage database~\cite{weng2019deepchain} or as a system for authorization and authentication~\cite{rathore2019blockdeepnet}. However, this shallow integration with blockchain fails to provide FL with significant improvements in terms of credibility and resilience against single points of failure. Besides, while the smart contract is designed to facilitate agreements among participants, its potential role in coordinating nodes to generate personalized models has not been thoroughly investigated.

To resolve the problems arising from personalized models and the shallow integration of blockchain, a smart contract-based edge intelligence framework for IoT systems (SCEI) is proposed in this paper. In SCEI, personalized models are trained based on the balanced weight of local and global models to ensure good performance on both local and skewed data. Specifically, a novel decentralized algorithm is developed on the smart contract to facilitate joint model training and dynamically adjust the weights according to local accuracies on edge nodes. In addition, a novel committee election consensus algorithm developed on the blockchain is fully utilized by the smart contract, retaining the decentralized nature while improving efficiency.

The main contributions of this paper are listed below.
\begin{itemize}
\item A smart-contract-based edge intelligence framework \emph{SCEI} is developed on a novel committee election consensus algorithm, which handles non-iid challenges while tackling credibility and security concerns for personalized model learning in IoT Systems.
\item A novel decentralized algorithm is developed on the smart contract for collaborative model training and dynamic weight adjusting according to local accuracies, which ensures optimal model accuracy for both local and skewed data.
\item An open-sourced prototype\footnote{The Github link is \url{https://github.com/xuchenhao001/EASC}.} is presented with extensive experiments conducted to show SCEI has improved model accuracy over benchmarks with acceptable overhead measured by time cost for each training round.
\end{itemize}

\section{Related Work}
The related work is introduced from three aspects, including FL and personalized learning in IoT, blockchain-empowered FL for edge intelligence, and smart contract driven FL.

\subsection{Federated Learning and Personalized Learning in IoT}

FL was first introduced in 2017~\cite{mcmahan2017communication} as a method to efficiently train machine learning models on edge devices~\cite{xu2021asynchronous, qu2022fl, xu2021bafl}. Most FL schemes aim to obtain an optimal global model by leveraging distributed local datasets across nodes. However, the performance variance of the global model on different nodes is inevitable~\cite{deng2020adaptive, hu2020personalized, mcmahan2021advances}, which has motivated recent research to focus on personalized model training. There are four main approaches to training personalized models: (1) localized independent training, where models are trained solely on local data~\cite{khodak2019adaptive}; (2) clustering training, where models are trained within clusters that exhibit similar data distributions~\cite{smith2017federated, shlezinger2020communication}; (3) optimizing personalized models through user's context~\cite{wang2019federated}; and (4) co-training of personalized models and global models~\cite{deng2020adaptive}. Our work aligns with approach (4), which aggregates weighted global and local models to get personalized models. However, most existing learning schemes lack a perception of local datasets and an optimization for skewed local data, except for the approach proposed by Deng \emph{et al.}~\cite{deng2020adaptive}. Their work leverages aggregated models to train personalized local models, which is similar to SCEI. However, their approach relied on a centralized aggregation server during training, whereas SCEI is fully distributed. Besides, SCEI incorporates the consideration of local accuracies from edge nodes to determine the optimal balancing weight in each training round, enhancing trustworthiness and practicality when dealing with skewed data. The scheme proposed by Deng \emph{et al.}~\cite{deng2020adaptive} serves as a benchmark in empirical studies.

\subsection{Blockchain Empowered Federated Learning for Edge Intelligence}

The integration of public (Proof-of-Work based) blockchain with FL was initially proposed in~\cite{kim2019blockchained}. So far, blockchain has been utilized to (1) select reliable participating nodes for FL~\cite{kang2019incentive}, (2) securely and reliably store local and global models~\cite{rathore2019blockdeepnet, warnat2021swarm}, (3) reduce the risk of a single point of failure from the centralized aggregate server~\cite{li2021blockchain, pokhrel2020federated, lu2020communication, ramanan2020baffle, xu2022efficient, xu2021bafl}, and (4) resist poisoning attacks by detecting and penalizing attackers~\cite{desai2021blockfla, xu2022efficient, xu2021bafl}. However, these lines of research focus solely on reaching an optimal global model accuracy without considering local model accuracy or data skew, limiting their application for most real-world scenarios (e.g., Smart Hospitals). Furthermore, the high computational load of blockchain reduces the feasibility of FL. To address these limitations, several asynchronous federated learning schemes have been proposed to improve efficiency by reducing the waiting time for aggregation~\cite{xu2021asynchronous}. However, due to the aggregation of outdated models from slow nodes, the improved efficiency usually comes at the expense of model accuracy, as evidenced by experimental results. This paper develops a decentralized algorithm executed on the smart contract to balance the weight of local and global models, leading to improved efficiency and performance when dealing with skewed data.

\subsection{Smart Contract Driven Federated Learning}

The smart contract, a distributed program in blockchain, possesses several features such as self-verification, self-execution, and tamper-resistance. These attributes enable the smart contract to execute various distributed operations on the blockchain at a minimal cost~\cite{mohanta2018overview}. The smart contract has been utilized to 1) incentive nodes to participate in FL~\cite{weng2019deepchain}, 2) record and verify cached data on edge devices~\cite{cui2020creat}, 3) manage edge training in IoT~\cite{rahman2020secure}, and 4) facilitate transaction authentication and user verification~\cite{nguyen2021federated, warnat2021swarm}. However, these lines of research have not fully explored the potential of the smart contract with the underlying consensus algorithm during the FL training process. In SCEI, blockchain and smart contracts are tightly integrated into the distributed learning process to achieve decentralized edge intelligence. This tight integration enables SCEI to strike a remarkable balance between model accuracy, scalability, credibility, and security.

\section{Proposed Solution}

SCEI is explained from four aspects, including the system architecture and workflow (Section~\ref{sec:system_architecture}), the smart contract coordinator and model weight balancing (Section~\ref{sec:weight_balancing}), the committee election and security (Section~\ref{sec:committee_election}), and the asynchronous solution (Section~\ref{sec:asynchronous_solution}). The relevant parameter symbols used in this section are listed in Table~\ref{table:symbols}.
\begin{table}[htp]
	\rowcolors{2}{gray!20}{white}
	\renewcommand{\arraystretch}{1.3}
	\caption{Symbols and Meanings}
	\label{table:symbols}
	\centering
	\begin{tabular}{p{0.12\linewidth}|p{0.78\linewidth}}
	\rowcolor{gray!50}
	\hline
	\textbf{Symbol} & \textbf{Meaning} \\
	\hline
	$k,~K$ & the node number and the total number of nodes\\
	$t,~T$ & the round number of federated optimization and its upper limit\\
	$r,~R$ & the negotiation round number and its upper limit\\
	$w_{G}$ & the global model\\
	$w_{k}$ & the local model of node $k$\\
	$w_{kr}$ & the personalized local model of node $k$ in negotiation round $r$\\
	$w_{kp}$ & the personalized local model of node $k$\\
	$\alpha$ & the local model weight\\
	$\alpha_l, \alpha_u$ & the lower and upper bounds of the local model weight\\
	$\alpha_{r*}^{t}$ & the optimal local model weight in training round $t$\\
	$A_{kr}$ & the test accuracy of node $k$ in negotiation round $r$\\
	$\bar{A}_{r}$ & the average test accuracy in negotiation round $r$\\
	$\gamma$ & the ratio of committee size to the number of nodes\\
	$C$ & the identity set of the committee \\
	$C_L^t$ & the committee leader identity in training round $t$ \\
	$H$ & the MD5 hash function\\
	\hline
	\end{tabular}
\end{table}

\subsection{System Architecture and Workflow}
\label{sec:system_architecture}

\begin{figure}[htp]
    \centering
    \begin{center}
    \includegraphics[width=\linewidth]{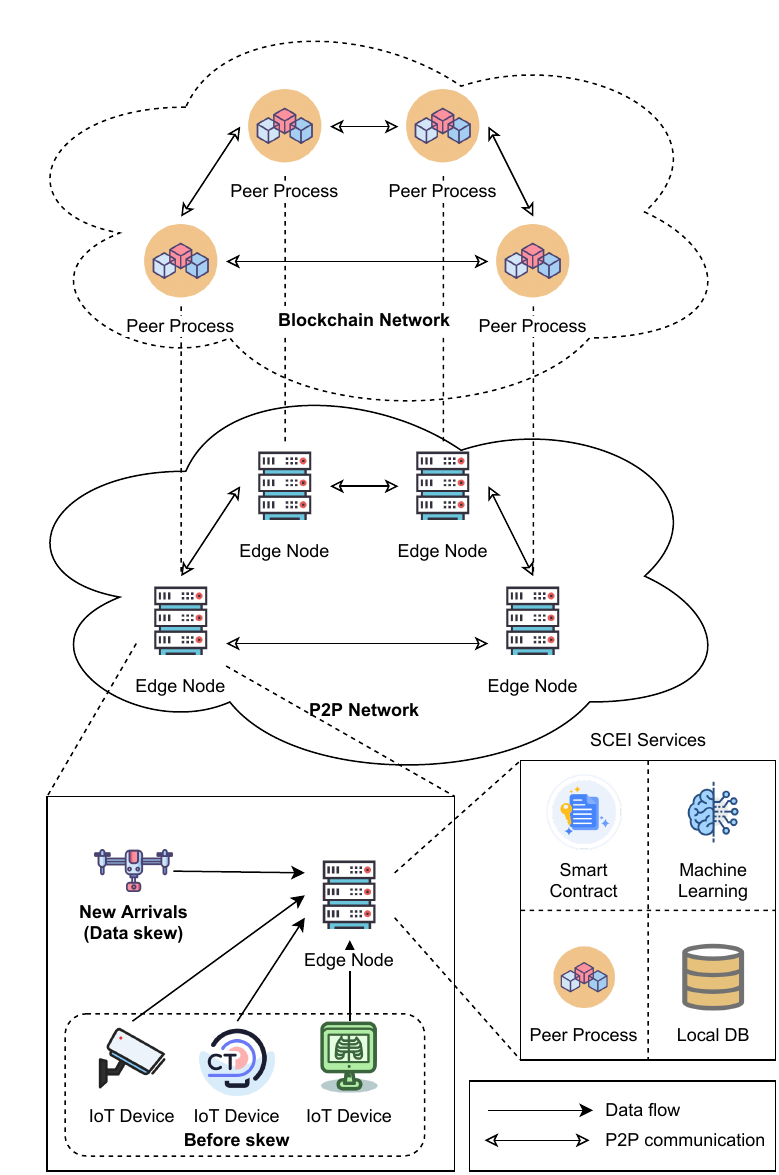}
    \end{center}
    \caption{The system architecture.}
    \label{fig:system-architecture}
\end{figure}

SCEI is built upon a consortium blockchain, with core component peers deployed on edge nodes, forming the backbone of the system. These peers execute the smart contract, maintain a shared ledger, and establish a peer-to-peer communication network among themselves, as shown in Fig.~\ref{fig:system-architecture}. To meet the requirements of storage capacity and computational power, the peers are strategically positioned on edge nodes and are responsible for collecting data from IoT devices within the same organization. For example, in a smart hospital system, peers deployed on edge nodes within medical imaging departments across hospitals join FL and collectively train models using healthcare data acquired from medical devices to categorize X-ray test results. Similarly, in a multi-farm smart farming system, peers running on edge nodes gather image data from IoT devices and utilize it to train models that predict the growth of specific crops.

Each edge node in the consortium blockchain undergoes verification of its identity by other nodes. A new edge node can only join the network if the majority of participants agree, which aligns with real-world scenarios. Upon joining SCEI, edge nodes are assigned sequential identities ranging from $1$ to $K$. Besides, in SCEI, the smart contracts running on edge nodes are uniform and govern the process of distributed model training. To improve scalability, all original models in SCEI are stored in the local database of the edge nodes, with only MD5 hash values uploaded to the shared ledger. For other nodes to access the original model, the owner of the model must generate a model download address and upload it to the shared ledger.

During each training round, a committee comprising a predetermined fraction ($\gamma$) of nodes is selected based on the hash value of the global model from the previous round. This committee consists of a leader and several members. Section~\ref{sec:committee_election} explains the process of committee election. The committee utilizes the Raft consensus algorithm~\cite{ongaro2014search} to synchronize the original models. Compared to Paxos~\cite{bolosky2011paxos} and other consensus algorithms, Raft has demonstrated superiority due to its simpler and more comprehensible processes, adequate completeness for practical system requirements, proven safety properties, and comparable efficiency~\cite{ongaro2014search}. Specifically, Raft employs a more effective form of leadership than other consensus algorithms by restricting the flow of log entries from the leader to other members. This simplifies the management of the replicated log and enhances the comprehensibility of Raft. Besides, Raft resolves conflicts simply and rapidly by electing leaders based on randomized timers, which introduces only a minimal additional mechanism to the heartbeats already required for other consensus algorithms. 

\begin{figure*}[ht]
    \centering
    \includegraphics[width=0.9\linewidth]{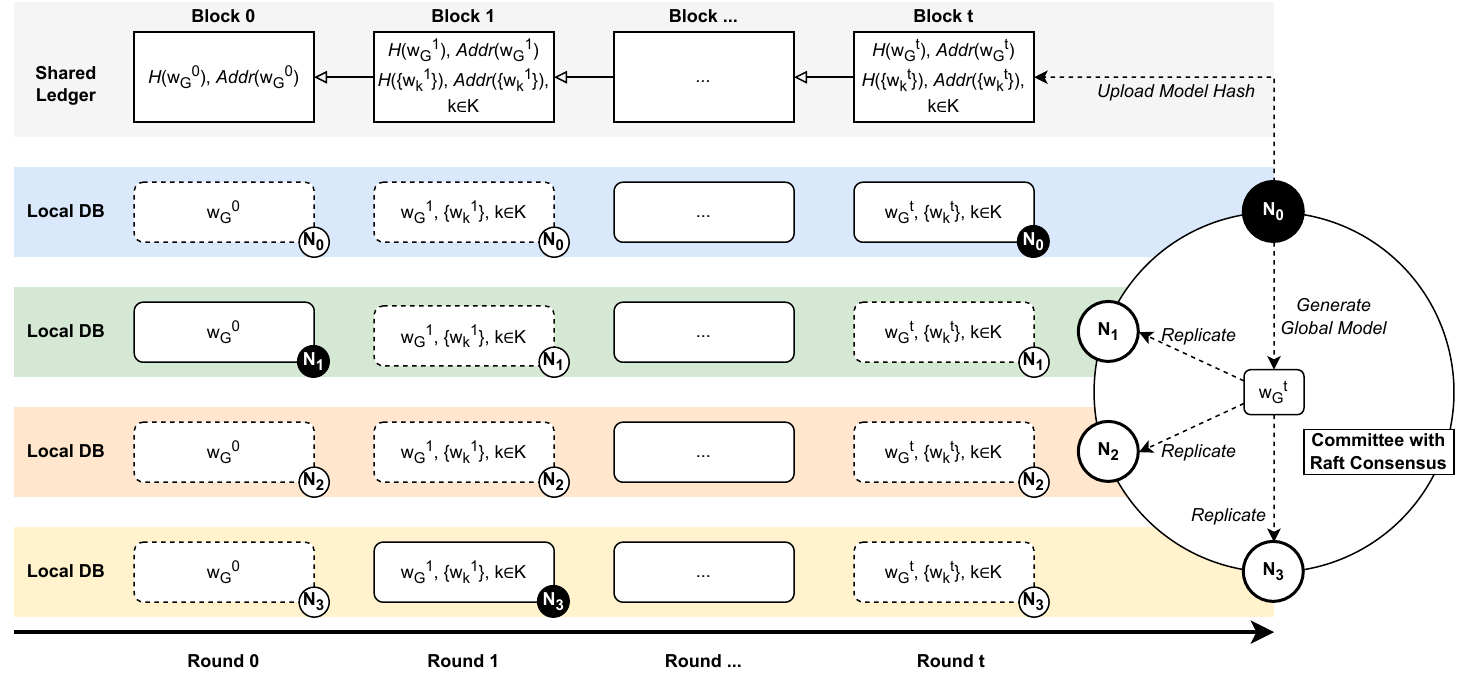}
    \caption{The committee structure leverages the Raft consensus to replicate the global model. The committee leader, depicted in black, assumes the role of generating the global model, distributing it to committee members, and uploading the corresponding hash value to the shared ledger in each round.}
    \label{fig:consensus}
\end{figure*}

Fig.~\ref{fig:consensus} demonstrates the committee with Raft consensus and the replication process of the global model in training round $t$. Assume there are four edge nodes ($N_0, \dots, N_3$) elected as the committee for training round $t$, where $N_0$ serves as the committee leader. Although not explicitly shown in the figure, the other edge nodes are connected to the committee. Each edge node possesses its own local database for storing the original model while maintaining the shared ledger of the blockchain.

The workflow initiates with the training of an initial global model $w_G^0$ on node $N_1$, after which the hash value of the initial global model $H(w_G^0)$ and the corresponding model download address $Addr(w_G^0)$ are uploaded to the shared ledger. Specifically, assuming the IP address of the first node is $\text{IP}_1$, $Addr(w_G^0)$ can be interpreted as $\text{http://IP}_1\text{/}H(w_G^0)$. Following this, the first training round commences.

In each training round, such as round $t$, all nodes retrieve the base model $w_G^{t-1}$ by using the download address provided in the shared ledger. Each node then utilizes its local data to train a new local model based on the base model. Afterward, the newly trained local model $w_k^t$ from node $k$ is uploaded to a randomly selected committee member, and replicated to all nodes within the committee by Raft. Concurrently, both the model download address $Addr(w_k^t)$ and the model hash value $H(w_k^t)$ are uploaded to the shared ledger, allowing other nodes to download and validate the original local model $w_k^t$. Upon receiving local models from all nodes, i.e. $\{w_k^t\}, k \in K$, the committee leader $N_0$ calculates a new global model $w_G^t$ using Eq.~\ref{eq:FedAvg}. The hash value $H(w_G^t)$ and download address $Addr(w_G^t)$ are then uploaded to the shared ledger. Although the committee leader generates $w_G^t$, all committee members are able to verify it based on their replicated local models. Thereafter, all nodes obtain $w_G^t$ using the download address provided in the shared ledger. 

\begin{gather}
\label{eq:FedAvg}
w^{t}_{G} = \frac{1}{K} \sum_{k=1}^{K}w_{k}^{t}
\end{gather}

Next, nodes generate their personalized local models by proportionally integrating the global and the local model with the help of the smart contract, as detailed in Section~\ref{sec:weight_balancing}. This integration allows the personalized local models to learn features from both the local and global models. With this step, training round $t$ is finalized. The entire training process finishes when the value of $t$ reaches $T$. 

Uploading hash values rather than original models to the shared ledger ensures the scalability of SCEI. In addition, the global models from earlier rounds are removable after a new global model has been successfully replicated to all nodes, which helps free up storage space and further improves scalability. 

\subsection{The Smart Contract Coordinator and Model Weight Balancing}
\label{sec:weight_balancing}

\begin{figure}[htp]
    \centering
    \includegraphics[width=\linewidth]{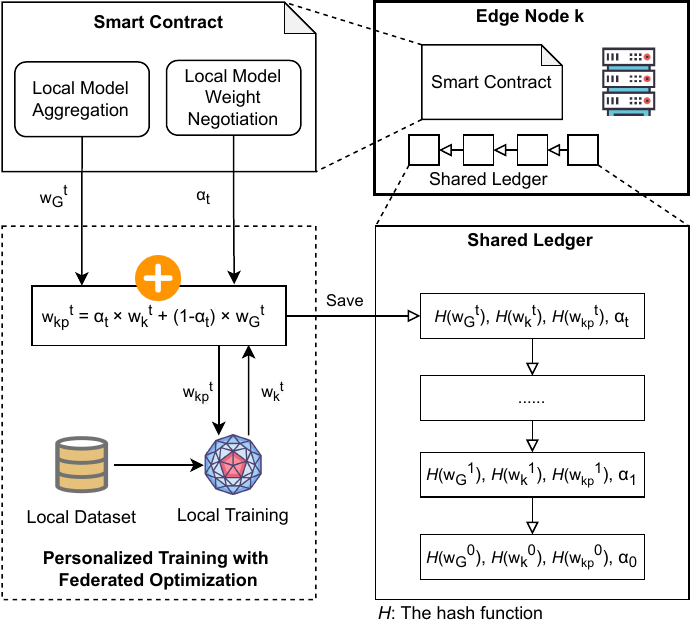}
    \caption{The integration of machine learning and smart contract.}
    \label{fig:integration-architecture}
\end{figure}

As shown in Fig.~\ref{fig:integration-architecture}, both the smart contract and the shared ledger are involved in the training process. The smart contract is responsible for 1) local model aggregation and 2) local model weight negotiation. Throughout the training process, the shared ledger records important information such as the hash values of models and the optimal local model weight.

To achieve improved personalized local models, a hyperparameter $\alpha$ that balances local and global models is introduced and is defined as Eq.~\ref{eq:scei}. In Eq.~\ref{eq:scei}, the variables represent the following: $t$ corresponds to the current training round, $w_k^t$ represents the local model on node $k$, $w_G^t$ denotes the global model obtained through FedAvg~\cite{mcmahan2017communication}, and $w_{kp}^{t}$ represents the personalized local model. The hyperparameter $\alpha$ plays a significant role in achieving a balance between the global and local models, particularly in handling non-iid datasets with the data distribution skew. When $\alpha=0.0$, the local model is disregarded, and SCEI is equivalent to FedAvg. Conversely, when $\alpha=1.0$, the personalized local model in SCEI becomes independent and loses its generalization capabilities.
\begin{gather}
\label{eq:scei}
w_{kp}^{t}=\alpha \times w_k^t+(1-\alpha) \times w_{G}^{t}.
\end{gather}

The optimal value of $\alpha$ cannot be predicted in advance due to the uncertainty of data distribution among the nodes~\cite{deng2020adaptive}. In SCEI, the smart contract enforces each node to test multiple $\alpha$ values during each training round to determine the best one. As shown in Algorithm~\ref{alg:scei}, the corporation of the smart contract with the training procedure on node $k$ is illustrated in lines $5$ to $11$ and $21$ to $30$. Specifically, in training round $t$, the smart contract uploads the hash value of the global model $H(w_G^t)$ to the shared ledger, triggering node $k$ to download the global model $w_G^t$ from the committee leader $C_L^t$, as indicated in lines $5$ and $21$. Suppose $R$ denotes the number of rounds to negotiate $\alpha$, $\alpha_l$ and $\alpha_u$ represent the lower and upper bounds of $\alpha$, and $step$ represents the increment of $\alpha$ in each negotiation round, which is calculated by Eq.~\ref{eq:step} and implemented in line $22$.
\begin{gather}
\label{eq:step}
step = \frac{\alpha_u - \alpha_l}{R}
\end{gather} 
Next, node $k$ generates multiple temporary personalized local models $w_{kr}^{t}$ with different $\alpha$ values for $R$ negotiation rounds and tests the accuracy of $w_{kr}^{t}$ using the local test dataset, as shown in lines $23$ to $27$. Following this, in line $28$, node $k$ uploads the local test accuracy set \{$A_{kr}^{t}$\} and the corresponding $\alpha$ set \{$\alpha_r$\} to the shared ledger, which signals the smart contract to stop waiting in line $6$. For each negotiation round, the smart contract averages $\{A_{kr}^{t}\}, k \in K$ and generates the average test accuracy $\bar{A}_{r}^{t}$, as illustrated in lines $7$ to $9$. The smart contract then employs one of two strategies to select the optimal $\alpha$: (A) maximizing the average of the test accuracies, or (B) minimizing the variance of the test accuracies. If Strategy A is chosen, the smart contract selects the negotiation round $r$ that yields the maximum $\bar{A}_{r}^{t}$, disregarding any relatively low test accuracies, as shown in lines $10$ to $11$. On the contrary, if Strategy B is adopted, the smart contract selects $r$ that exhibits the minimum variance of $\{A_{kr}^{t}\}, k \in K$. Subsequently, the optimal local model weight $\alpha_{r*}$ is determined and uploaded to the shared ledger. As a result, node $k$ calculates the optimal personalized local model $w_{kp}^t$ for training round $t$, as shown in lines $29$-$30$. If the required number of optimization rounds is not reached, node $k$ initiates a new round of training based on $w_{kp}^t$, as shown in line $16$.

Strategy A is tailored for a static personalized objective, where each node possesses a fixed personalized learning objective. For example, in a smart farming system, the personalized learning goal of one farm is to predict the growth of wheat, whereas the personalized learning goal of another farm is to predict the growth of beans~\cite{idoje2021survey}. Strategy B caters to a dynamic personalized objective, where nodes continuously learn and adapt to new data arrivals. For example, the learning goal of a smart hospital system is to recognize COVID-19 cases, but it changes over time as new viral strains emerge~\cite{korber2020tracking}. Therefore, the personalized learning objective of the hospital shifts, leading to dynamic updates in the personalized model based on gradients learned from models in other hospitals that have detected the new variants.

\begin{algorithm}[ht]
	\caption{SCEI}
	\label{alg:scei}
	\begin{algorithmic}[1]
	\Procedure{SmartContract}{}
	    \State initialize the global model $w_{G}^{0}$
	    \State wait for local updates \{$w_{k}^{t}$\}, $k \in K$
	    \State $w_{G}^{t} \leftarrow $ FedAvg(\{$w_{k}^{t} $\})
	    \State upload $H(w_G^t)$ to the shared ledger
	    \State wait for \{$A_{kr}^{t}$\} and \{$\alpha_{r}$\} ($r \in R$, $k \in K$)
	    \For {$r=1,2, \dots ,R$} \textit{ // R rounds of negotiation}
	        \State $\bar{A}_r^{t} \leftarrow \sum_{k=1}^{K} A_{kr}^{t} / K$
	    \EndFor
	    \State $ r* \leftarrow \argmax_{r \in R}(\{\bar{A}_{r}^{t}\})$
	    \State upload $\alpha_{r*}$ to the shared ledger
	\EndProcedure
	\\
	\Procedure{LocalTraining}{$k$} \Comment{\emph{Run Parallelly}}
	    \State $w_{kp}^{0} \leftarrow w_{G}^{0}$ \textit{ // initiate the personalized local model}
	    \For {$t=1,2, \dots ,T$} \textit{ // T rounds of optimization}
	        \State $C_L^t \equiv H(w_G^{t-1}) \Mod K$
	        \State $C^t \leftarrow \{C_L^t, C_L^t+1, \ldots, C_L^t + [\gamma K]\} $
	        \State $w_{k}^{t} \leftarrow $ LocalTrain($w_{kp}^{t-1}$, localTrainDataset)
	        \State upload $H(w_k^t)$ to the shared ledger
	        \State download $w_{G}^{t}$ from the committee leader $C_L^t$
	        \State $step \leftarrow (\alpha_u - \alpha_l) / R$
	        \For {$r=1,2, \dots, R$} \textit{ // R rounds of negotiation}
	            \State $\alpha_{r} \leftarrow \alpha_l + r * step$
	            \State $w_{kr}^{t} \leftarrow \alpha_{r} \times w_{k}^{t} + (1-\alpha_{r}) \times w_{G}^{t}$
	            \State $A_{kr}^{t} \leftarrow $ LocalTest($w_{kr}^{t}$ , localTestDataset)
	        \EndFor
	        \State upload \{$A_{kr}^{t}$\}, \{$\alpha_{r}$\} ($r \in R$) to the shared ledger
	        \State download $\alpha_{r*}$ from the shared ledger
	        \State $w_{kp}^{t} \leftarrow \alpha_{r*} \times w_{k}^{t} + (1-\alpha_{r*}) \times w_{G}^{t}$
	    \EndFor
	\EndProcedure
	\end{algorithmic}
\end{algorithm}

\subsection{Committee Election and Security}
\label{sec:committee_election}

The committee leader is responsible for aggregating local models and generating the global model in each training round, while the committee members are tasked with verifying the aggregation results. Ensuring the security of SCEI requires that the committee for each training round is distinct and unpredictable. Therefore, a new committee is elected for each round, based on the MD5 hash value of the most recent global model generated in the preceding round.

Specifically, the MD5 hash function denoted by $H$ is utilized to compute the hash value of base64-encoded models. Suppose $t$ represents the current training round number, and $k$ represents the number of nodes in SCEI. The hash value of the most recent global model generated in the preceding training round is $H(w_G^{t-1})$. As shown in lines $17$ to $18$ of Algorithm~\ref{alg:scei}, the identity of the committee leader, denoted as $C_L^t$, is determined by Eq.~\ref{eq:leader}.
\begin{gather}
\label{eq:leader}
C_L^t \equiv H(w_G^{t-1}) \Mod K.
\end{gather}
Assuming $\gamma$ represents the ratio of the number of nodes in the committee to the total number of nodes in SCEI, the identity set of committee members $C^t$ in training round $t$ is defined in Eq.~\ref{eq:committee}.
\begin{gather}
\label{eq:committee}
\{C_L^t+1,C_L^t+2,\ldots,C_L^t+[\gamma K]\}.
\end{gather}

The identification of the committee leader is verifiable for all nodes due to the unified global model in each training round. Besides, the unpredictability of the hash value for the latest global model ensures the randomness and fairness of the committee election process.

Assume there are five peers in SCEI to enable collaborative model training and ensure a fair committee election. The majority of these edge nodes are assumed to be honest, ensuring training the global model in the right direction and deterring malicious edge nodes from joining the network. Additionally, DDoS attacks are launched from external servers. In this context, the committee, randomly elected based on hash values of global models, offers resilience against a single point of failure and provides defense against DDoS attacks toward the centralized aggregate server. Furthermore, the immutable hash values on the blockchain help trace and verify malicious local models, preventing any evildoing by edge nodes. As a result, the security of the training process in SCEI is assured.

By leveraging the smart contract and blockchain to facilitate the model training process, SCEI ensures the trustworthiness of edge devices. The smart contract functions as a program, exhibiting self-verifiable, self-enforcing, and tamper-proof features, and is executed simultaneously across the nodes. Through a consensus among all nodes prior to deployment, the behavior of each edge device becomes predictable and aligned with the shared commercial objective. Moreover, the availability of models, accompanied by their verifiable hash values on the blockchain, enables the detection and blacklisting of any edge node attempting to upload malicious models.

\subsection{Asynchronous Solution}
\label{sec:asynchronous_solution}

To improve scalability, an asynchronous variant of SCEI (SCEI-Async) is introduced, which performs aggregation promptly upon receiving a new local model. The SCEI-Async algorithm closely resembles SCEI, except for the modifications in the smart contract, as depicted in Algorithm~\ref{alg:scei_async}. In this case, $t$ represents the version of the global model instead of the round number of federated optimization, and $t*$ refers to the most recent training round of node k, which is independent of $t$.

Upon receiving the latest local updates $w_{k}^{t*}$ from node $k$, a new global model $w_{G}^{t}$ is created by averaging $w_G^{t-1}$ and $w_{k}^{t*}$, as shown in lines $3$ and $4$. The hash value of $w_{G}^{t}$ is then uploaded to the shared ledger in line $5$. Subsequently, node $k$ tests various personalized local models using different $\alpha$ values $\{\alpha_{r}\}$, obtaining the corresponding local test accuracies $\{A_{kr}^{t*}\}$. In line $6$, node $k$ uploads the latest local test accuracy set $\{A_{kr}^{t*}\}$ and its corresponding $\alpha$ set $\{\alpha_{r}\}$ to the blockchain. As a result of the asynchronous aggregation strategy, the smart contract no longer waits for other nodes to upload their updated local test accuracies. Instead, the smart contract computes the average test accuracy $\bar{A}_r^{t}$ based on the most recent local test accuracies of other nodes stored in the blockchain, and selects the optimal negotiation round number $r*$, as shown in lines $7$-$11$.

In SCEI-Async, nodes are not required to wait for the completion of other nodes' training rounds. However, this results in a higher frequency of model aggregation, which introduces additional computing overhead than SCEI. Furthermore, the increased frequency of committee elections raises the risk of inconsistencies in the blockchain consensus. Additionally, since there is no synchronized training round in SCEI-Async, the local test accuracy set and the corresponding $\alpha$ set stored on the blockchain may not be up-to-date. As the value of $\alpha$ relies on the latest local test accuracy set stored in the blockchain, the negotiated $\alpha$ could be biased, causing a degradation of the accuracy of personalized models.

\begin{algorithm}[ht]
	\caption{SCEI-Async}
	\label{alg:scei_async}
	\begin{algorithmic}[1]
	\Procedure{SmartContract}{}
	    \State initialize the global model $w_{G}^{0}$
	    \State wait for the latest local updates $w_{k}^{t*}$,~$k \in K$
	    \State $w_{G}^{t} \leftarrow $ FedAvg($w_G^{t-1}, w_{k}^{t*}$)
	    \State upload $H(w_G^{t})$ to the shared ledger
	    \State wait for $\{A_{kr}^{t*}\}$ and $\{\alpha_{r}\}$ ($r \in R$) from node $k$
	    \For {$r=1,2, \dots ,R$} \textit{ // R rounds of negotiation}
	        \State $\bar{A}_r^{t} \leftarrow \sum_{k=1}^{K} A_{kr}^{t*} / K$
	    \EndFor
	    \State $ r* \leftarrow \argmax_{r \in R}(\{\bar{A}_{r}^{t}\})$
	    \State upload $\alpha_{r*}$ to the shared ledger
	\EndProcedure
	\end{algorithmic}
\end{algorithm}

In the experiments, SCEI-Async is utilized as a benchmark to showcase the advantages and disadvantages of utilizing an asynchronous aggregation strategy in SCEI.

\section{System Evaluation}
The experiments are conducted to answer the following research questions and assess the performance of \emph{SCEI}.
\begin{itemize}
  \item \textbf{RQ1:} Can \emph{SCEI} enhance model accuracy compared to the state-of-the-art approaches?
  \item \textbf{RQ2:} What is the impact of the number of nodes in the underlying network on model accuracy and convergence in \emph{SCEI}?
  \item \textbf{RQ3:} What is the computational and communication overhead associated with \emph{SCEI}?
\end{itemize}

\subsection{Experiment Setup}
The simulation environment consists of ten default nodes, each of which is a virtual machine (VM) with identical hardware and software configurations. Each VM is equipped with eight Cores, 8GB of RAM, and an NVIDIA GeForce GTX 1080 Ti GPU. In terms of software, Python 3.6 and PyTorch v1.6.0 are installed on each VM to support model training. The blockchain infrastructure relies on Hyperledger Fabric v2.2.0, while a middleware application is developed on Express.js v4.16.1 and deployed on each VM to facilitate peer communication within the blockchain network. Besides, the election timeout of Raft~\cite{ongaro2014search} is randomized within a range of $1$ to $2$ seconds. The default settings for the blockchain network are provided in Table~\ref{table:blockchain-parameter-setting}. 

\begin{table}[ht]
    \rowcolors{2}{gray!20}{white}
    \renewcommand{\arraystretch}{1.3}
    \caption{Parameter Settings for Blockchain}
    \label{table:blockchain-parameter-setting}
    \centering
    \begin{tabular}{c|c}
    \rowcolor{gray!50}
    \hline
    \textbf{Parameter} & \textbf{Value}\\
    \hline
    Block generation frequency
    & 0.2s\\
    The maximum number of messages in a block
    & 500\\
    The maximum size of a block
    & 100MB\\
    Election timeout in Raft
    & 1-2s\\
    \hline
    \end{tabular}
\end{table}

The performance of \emph{SCEI} is compared with three other state-of-the-art schemes: FedAvg~\cite{mcmahan2017communication}, Local Training~\cite{mcmahan2017communication} (each node trains its deep learning model locally and independently), and APFL~\cite{deng2020adaptive}. In line with~\cite{mcmahan2017communication,li2020privacy}, the parameter settings for personalized federated learning are described in Table~\ref{table:fl-parameter-setting}. To ensure generalization capability of neural networks and improve experimental efficiency, the range of $0.5$ to $0.8$ is selected for \emph{SCEI} to dynamically adjust the value of $\alpha$. 

\begin{table}[ht]
    \rowcolors{2}{gray!20}{white}
    \renewcommand{\arraystretch}{1.3}
    \caption{Parameter Settings for Personalized Federated Learning}
    \label{table:fl-parameter-setting}
    \centering
    \begin{tabular}{c|c}
    \rowcolor{gray!50}
    \hline
    \textbf{Parameter} & \textbf{Value}\\
    \hline
    The number of nodes $K$ & 10\\
    Local minibatch size $B$ & 10\\
    The local training rounds $E$ & 5\\
    The federated optimization rounds $T$ & 50\\
    The negotiation rounds $R$ & 10\\
    The proportion of committee nodes $\gamma$ & 0.3 \\
    Learning rate $\eta$ & 0.01\\
    The bounds of the local model weight $\alpha_l$, $\alpha_u$ & [0.5, 0.8]\\
    \hline
    \end{tabular}
\end{table}

Six public datasets are used in the experiments, including MNIST, CIFAR-10, CIFAR-100, IMAGENET~\cite{deng2009imagenet}, UCI~\cite{reyes2012human}, and REALWORLD~\cite{sztyler2016body}. MNIST, CIFAR-10, CIFAR-100, and IMAGENET are usually adopted as benchmark datasets~\cite{mcmahan2017communication, li2020privacy, xu2021asynchronous}. UCI and REALWORLD are human activity recognition datasets built from the recordings of $30$ participants performing six activities and $15$ participants performing eight activities, respectively.

To enable training of diverse personalized local models, the dataset settings are designed to be non-iid across nodes. Specifically, on each node, the dataset is sampled from four randomly selected classes within the original dataset. By default, each class contains $150$ image samples, except for UCI and REALWORLD where each class comprises $500$ image samples to ensure the model accuracy. To evaluate the performance of each personalized local model, a small portion of skewed data is introduced into the test dataset of each node. The skewed data is randomly sampled from classes other than the four selected classes. For example, in the case of the MNIST dataset, the training dataset on a node consists of images labeled ``0'', ``1'', ``2'', and ``3'', while the skewed test data is sampled from images labeled ``4'' to ``9''. The model accuracy is evaluated using the test dataset with varying data skew ratios, namely $0\%$ (no skew), $5\%$, $10\%$, $15\%$, and $20\%$. This indicates that the test data derived from images labeled ``4'' to ``9'' accounts for $0\%$, $5\%$, $10\%$, $15\%$, and $20\%$ of the overall test data, respectively.

In line with~\cite{mcmahan2017communication}, the experiments employ three training models: MLP, a multilayer-perceptron with ReLU activation, consisting of $2$ hidden layers and $200$ units; CNN, a deep neural network with two $5\times5$ convolution layers, a fully connected layer comprising $512$ units with ReLU activation, and a softmax output layer; RESNET, a CNN network with nine layers, featuring two residual blocks that allow the input to bypass certain layers~\cite{rajagopalan2022deep}.

To answer \emph{RQ1}, we compare the average local test accuracy of models using dynamic $\alpha$ and static $\alpha$ (ranging from $0.0$ to $1.0$). The average local test accuracy is computed by averaging the accuracies of personalized local models tested on each node, which quantifies the effectiveness of different schemes in training optimal local models. Then, the average local test accuracy of \emph{SCEI} is compared with that of Local Training, FedAvg, and APFL over $50$ rounds without data skew. Besides, the average local test accuracy is evaluated for different levels of data skew with the results presented in box plots.
To answer \emph{RQ2}, the average local test accuracy of models in \emph{SCEI} is compared with $5$, $10$, and $20$ nodes.
To answer \emph{RQ3}, the average overall time cost and the average communication time cost are recorded to reveal the computation and communication overheads of \emph{SCEI} over $50$ training rounds. The overall time cost of a training round is the time taken by a node to complete a training round, including communication time. The communication time cost is the time taken by a node to communicate with others, including committee election, model transmission, and uploading hash values to the blockchain.

\subsection{Results Analysis}

\subsubsection{RQ1. Model Accuracy}
\begin{figure*}
  \begin{center}
    \includegraphics[width=0.8\textwidth]{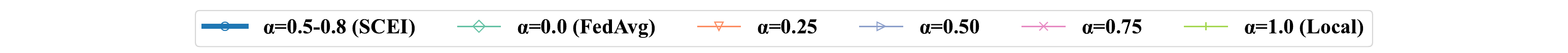}
    \subfigure[CNN on CIFAR-10]{
      \includegraphics[width=0.24\textwidth]{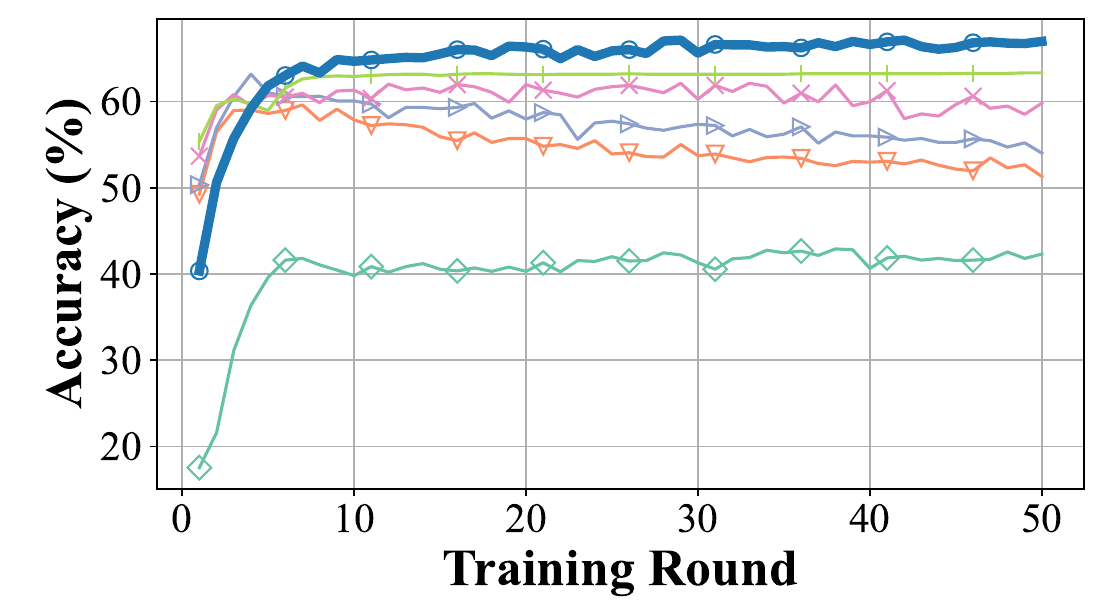}}
    \subfigure[RESNET on CIFAR-10]{
      \includegraphics[width=0.24\textwidth]{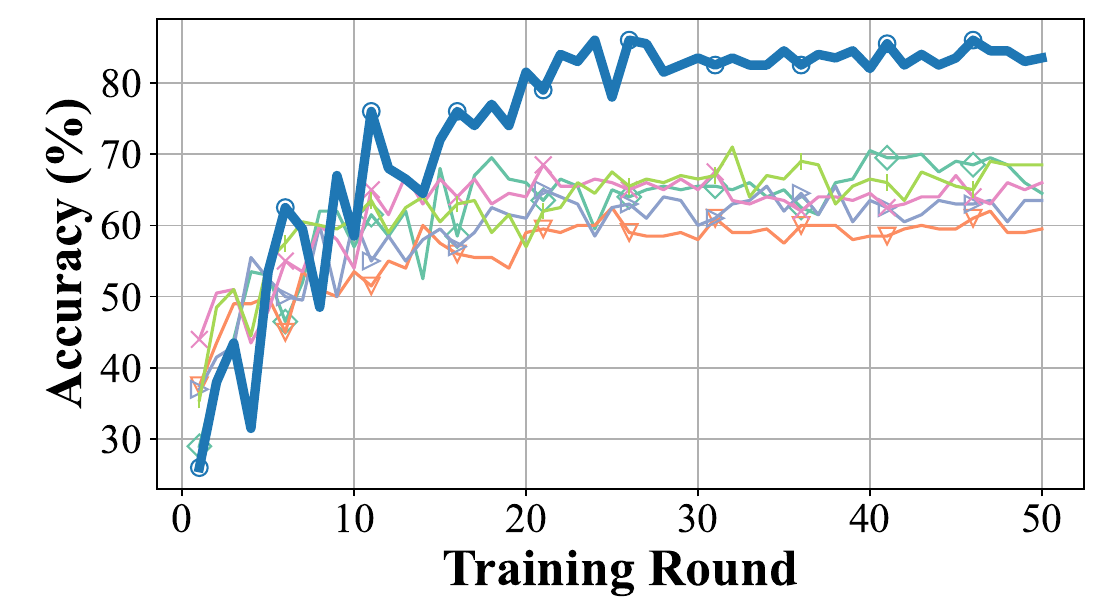}}
    \subfigure[CNN on MNIST]{
      \includegraphics[width=0.24\textwidth]{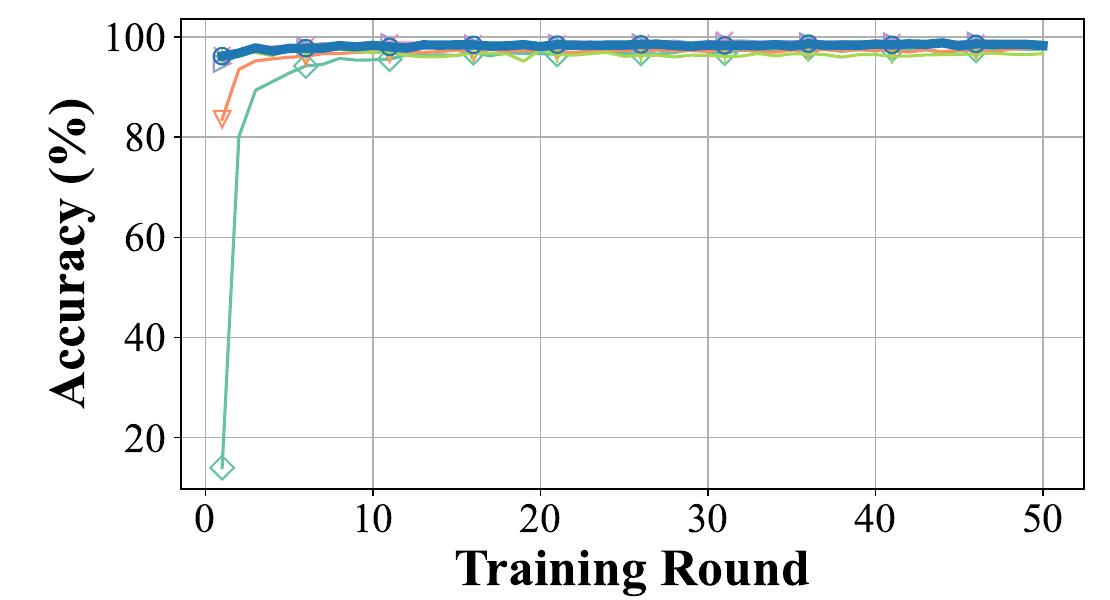}}
    \subfigure[MLP on MNIST]{
      \includegraphics[width=0.24\textwidth]{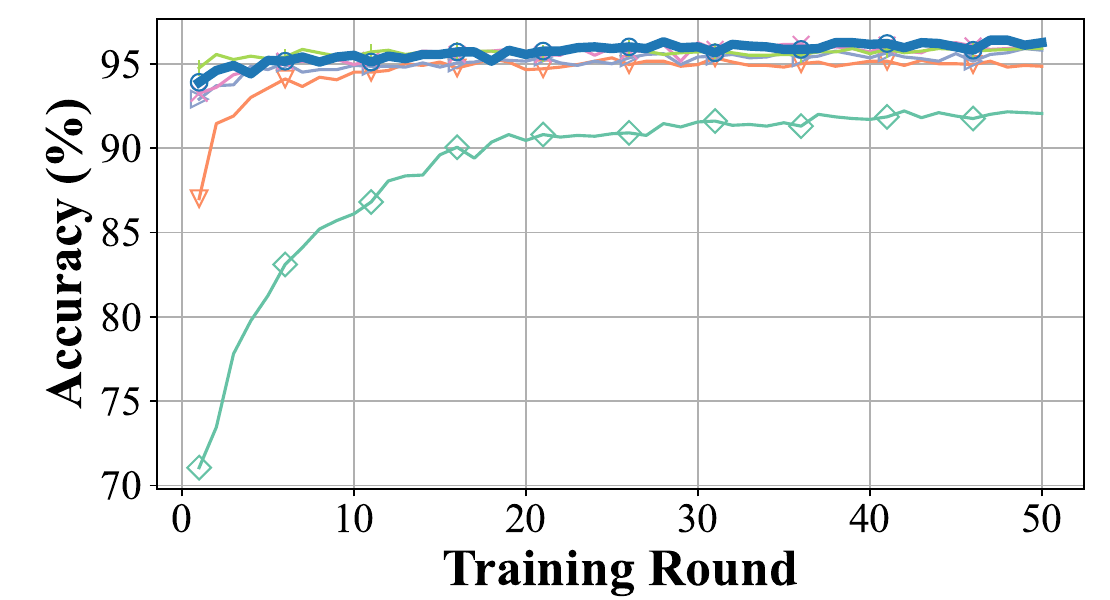}}
    \subfigure[CNN on CIFAR-100]{
      \includegraphics[width=0.24\textwidth]{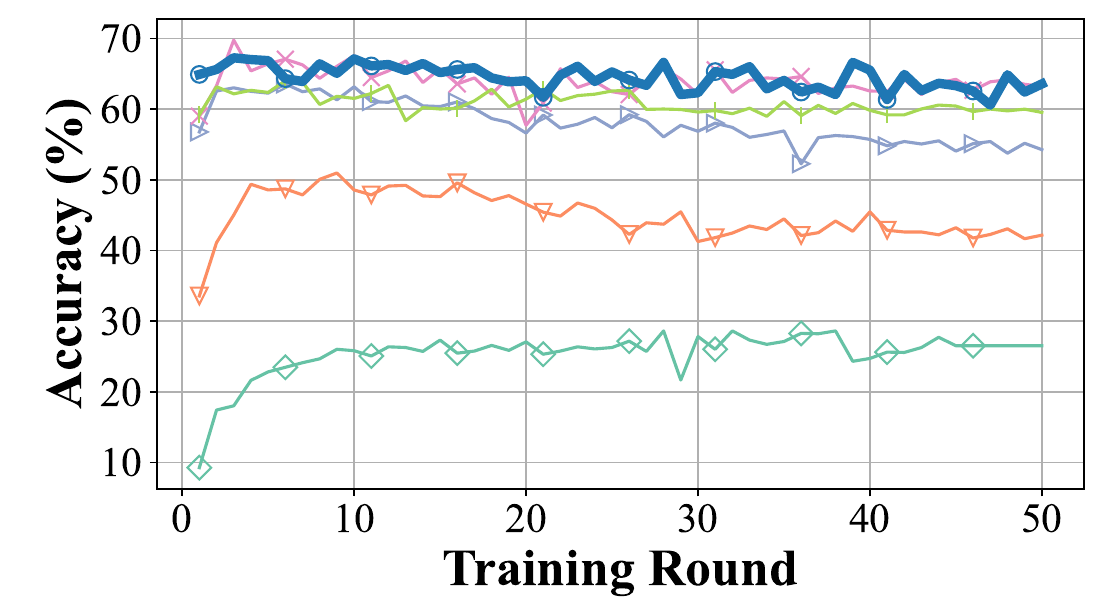}}
    \subfigure[CNN on UCI]{
      \includegraphics[width=0.24\textwidth]{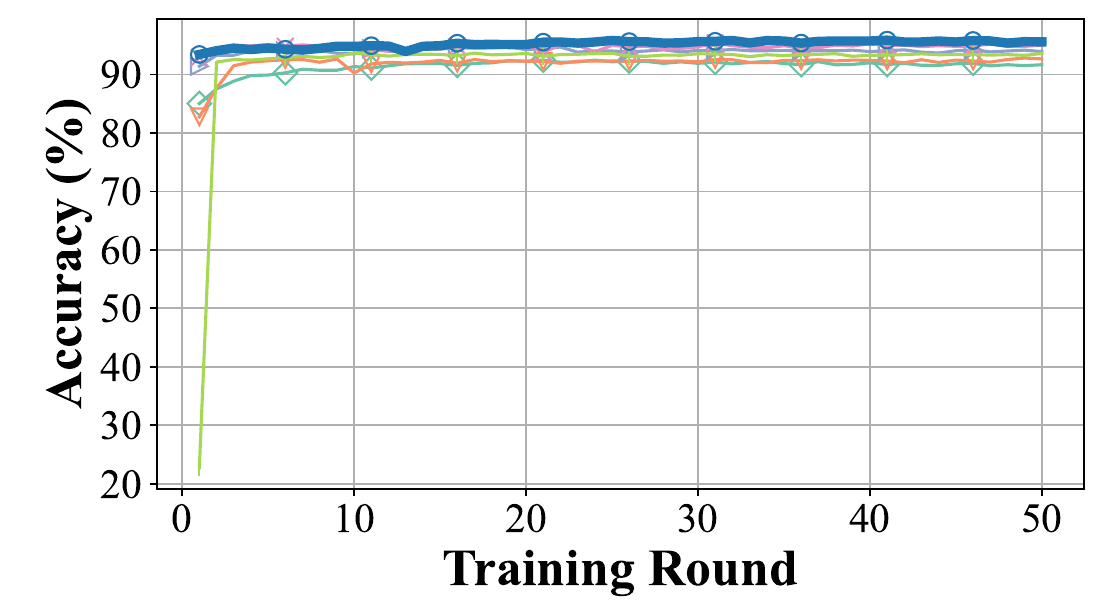}}
    \subfigure[CNN on REALWORLD]{
      \includegraphics[width=0.24\textwidth]{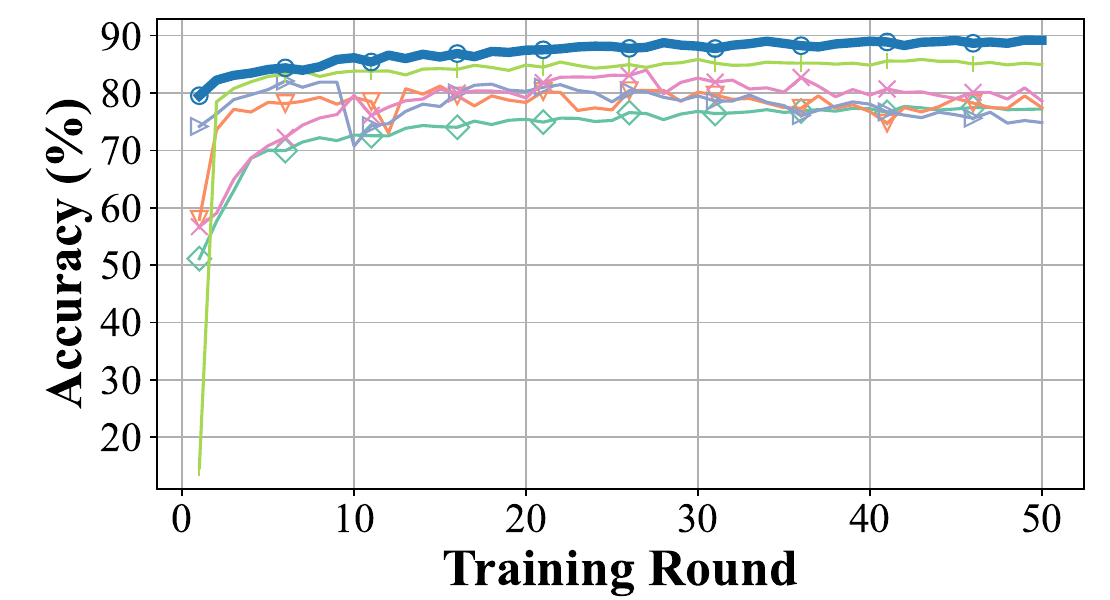}}
    \subfigure[CNN on IMAGENET]{
      \includegraphics[width=0.24\textwidth]{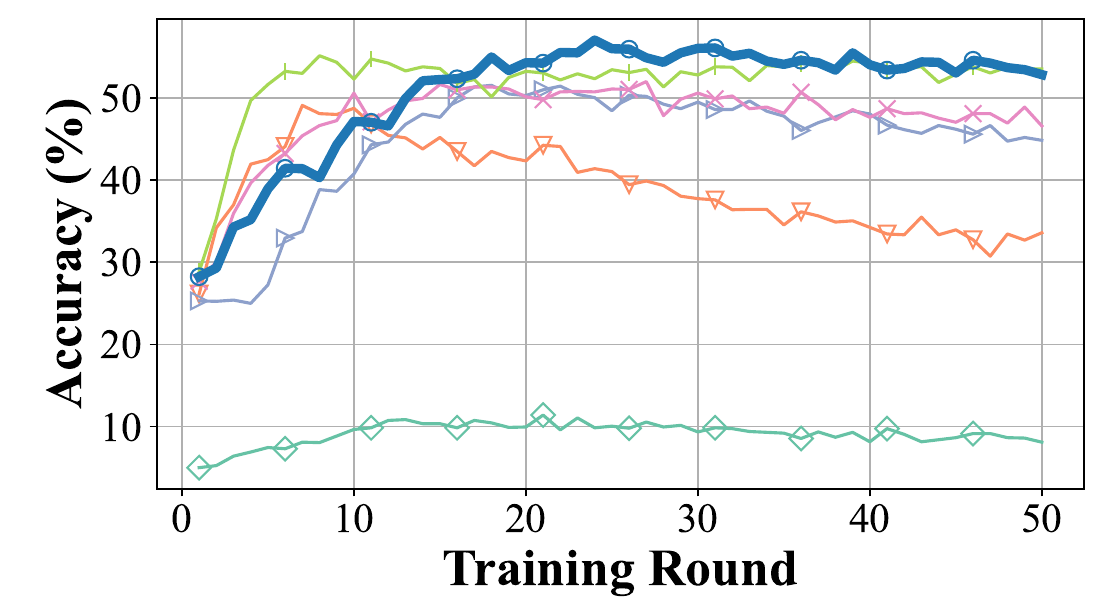}}
  \end{center}
  \caption{Comparison of average local test accuracy between SCEI with negotiated $\alpha$ values and SCEI with static $\alpha$ values.}
  \label{fig:acc_alpha}
\end{figure*}

When $\alpha=0.0$, \emph{SCEI} performs equivalently to FedAvg. Conversely, when $\alpha=1.0$, the personalized models trained by \emph{SCEI} lose their generalization capability and become similar to Local Training. Fig.~\ref{fig:acc_alpha} illustrates that the personalized models generated through the negotiation of $\alpha$ values achieve satisfactory accuracy after approximately $20$ training rounds, with \emph{SCEI} demonstrating a faster convergence speed than FedAvg.

\begin{figure*}
  \begin{center}
  \includegraphics[width=0.8\textwidth]{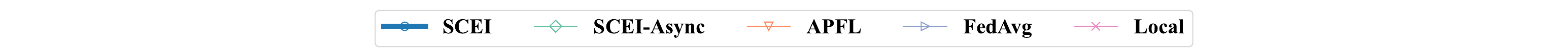}
    \subfigure[CNN on CIFAR-10]{
      \includegraphics[width=0.24\textwidth]{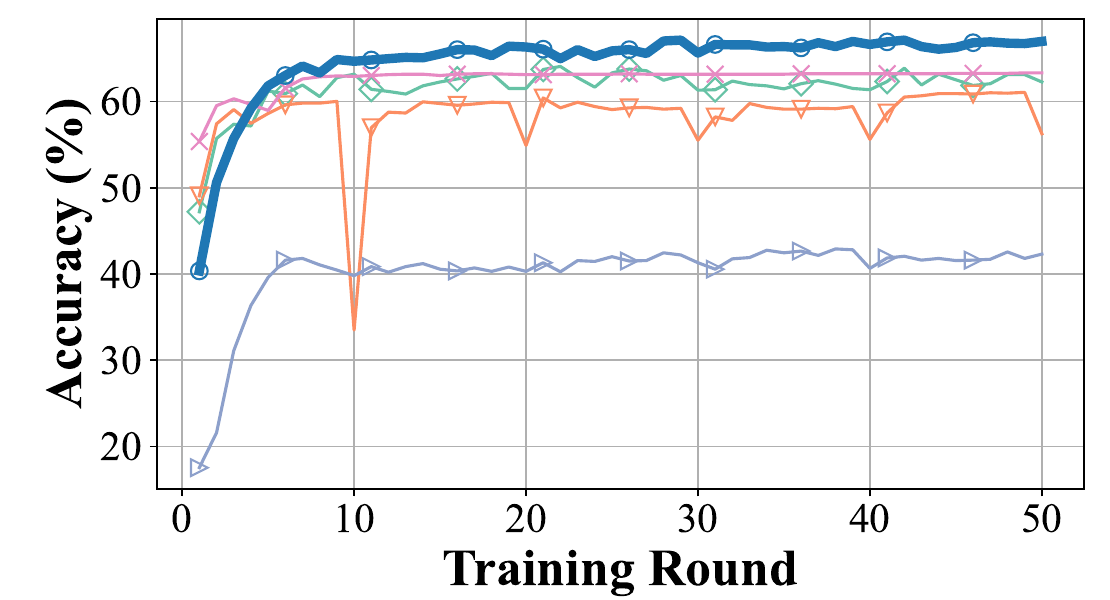}}
    \subfigure[RESNET on CIFAR-10]{
      \includegraphics[width=0.24\textwidth]{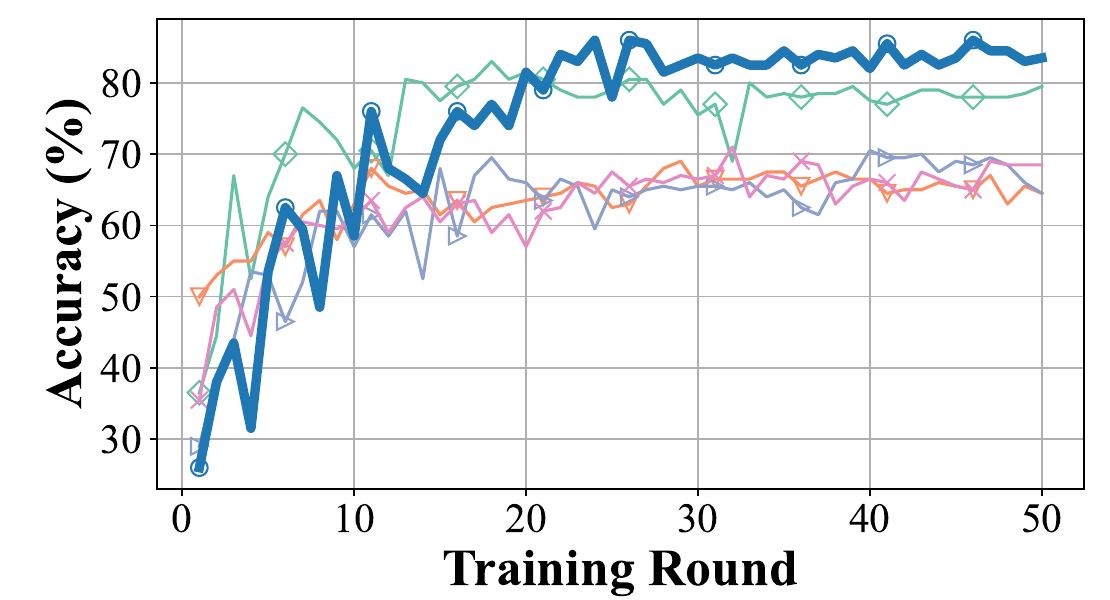}}
    \subfigure[CNN on MNIST]{
      \includegraphics[width=0.24\textwidth]{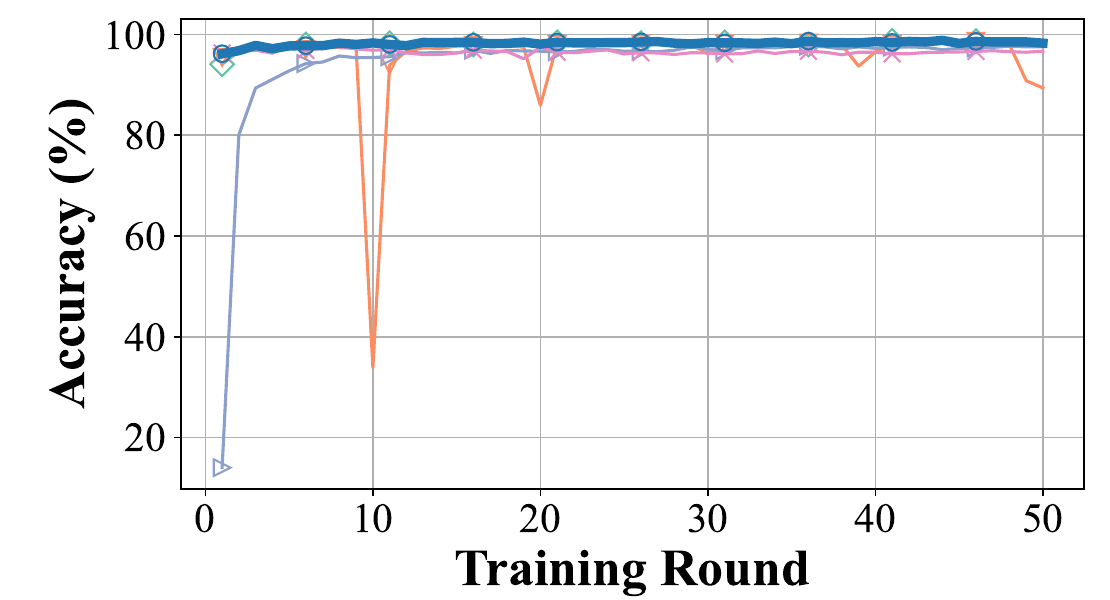}}
    \subfigure[MLP on MNIST]{
      \includegraphics[width=0.24\textwidth]{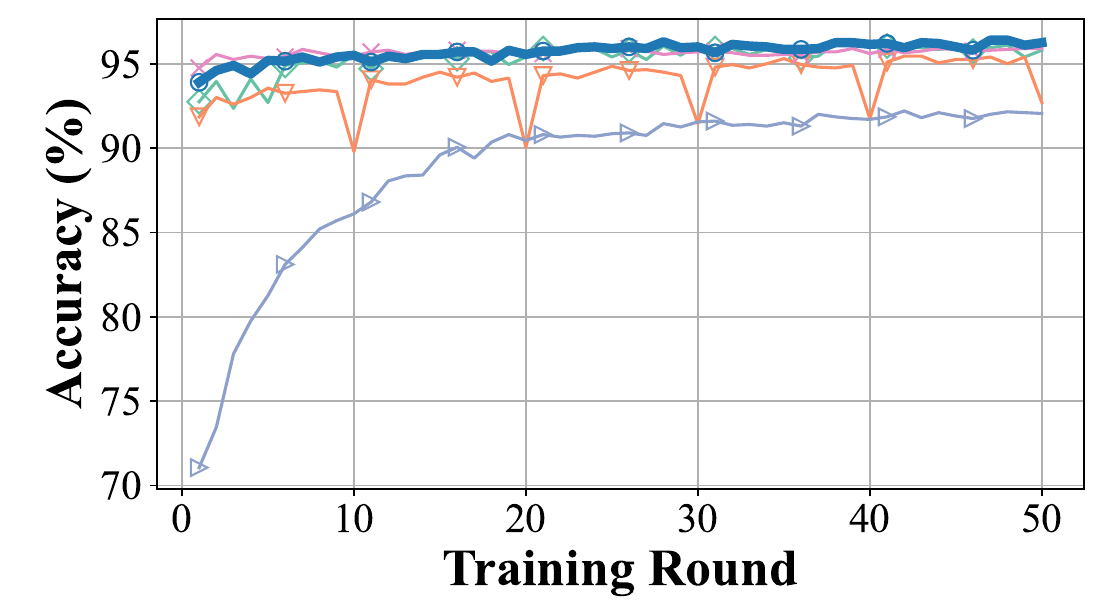}}
    \subfigure[CNN on CIFAR-100]{
      \includegraphics[width=0.24\textwidth]{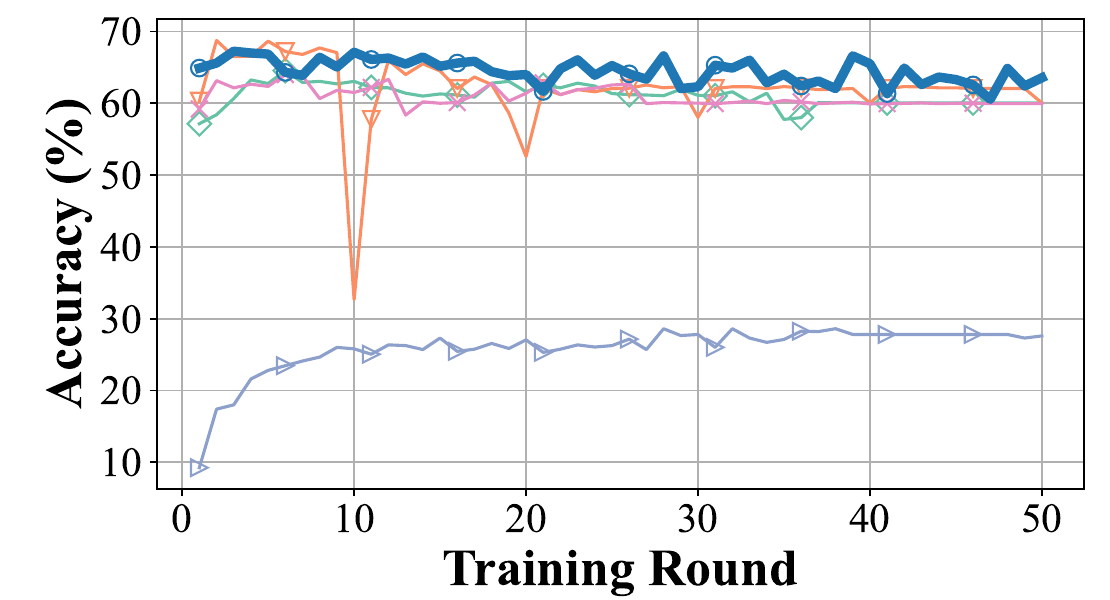}}
    \subfigure[CNN on UCI]{
      \includegraphics[width=0.24\textwidth]{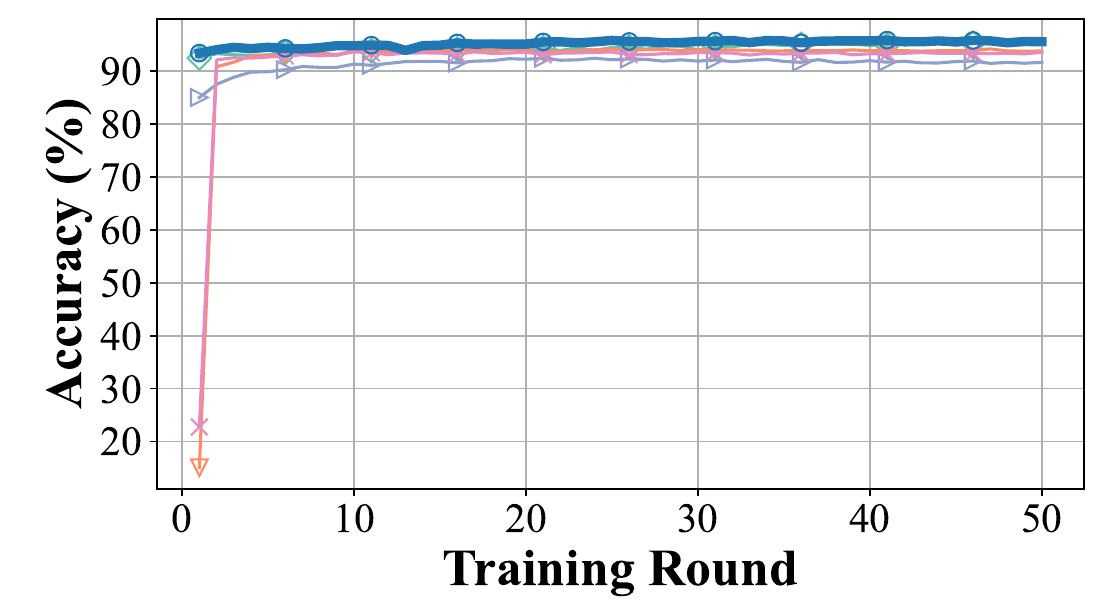}}
    \subfigure[CNN on REALWORLD]{
      \includegraphics[width=0.24\textwidth]{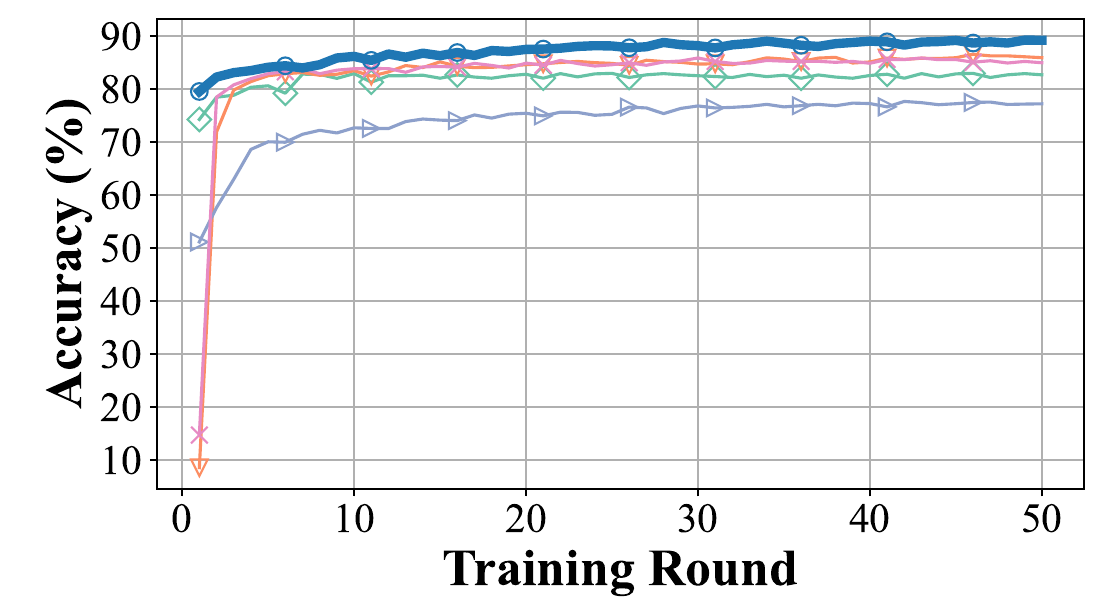}}
    \subfigure[CNN on IMAGENET]{
      \includegraphics[width=0.24\textwidth]{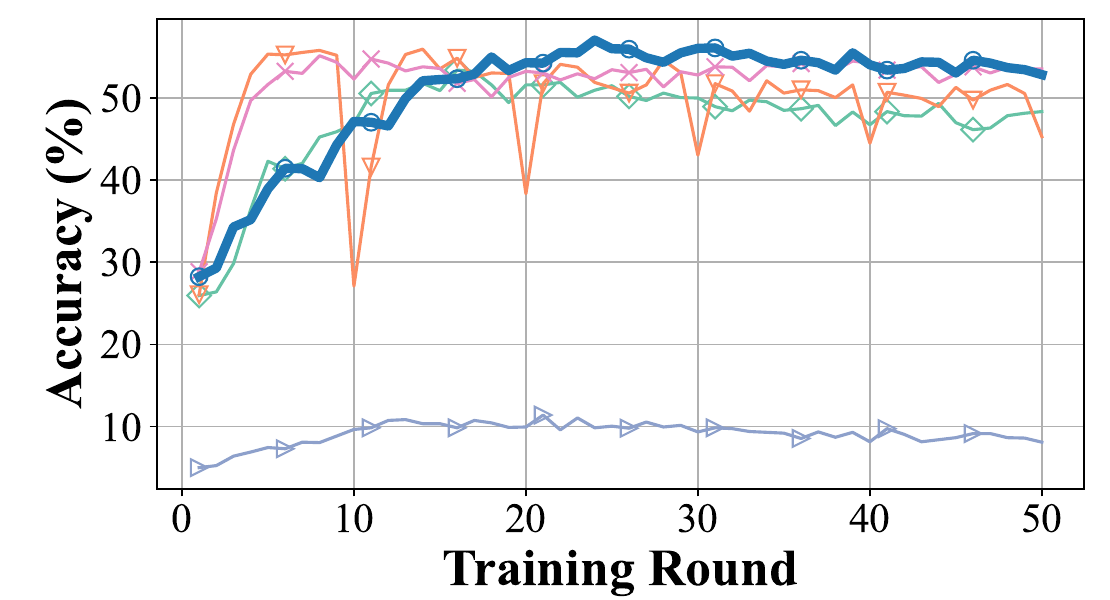}}
  \end{center}
  \caption{Comparison of average local test accuracy between SCEI and other schemes: FedAvg, APFL, and Local Training.}
  \label{fig:acc_sota}
\end{figure*}

Models trained with various static $\alpha$ settings exhibit varied performance, but the dynamically negotiated $\alpha$ consistently achieves the highest average local test accuracy. This is because \emph{SCEI} learns additional features from the global model, surpassing the performance of local training (when $\alpha>0.5$), and dynamically balancing the weights of local and global models based on the local test accuracy of each node. In contrast, FedAvg disregards personalized features and exhibits the worst performance, especially when training CNN on CIFAR-100 and IMAGENET datasets.

The comparison of average local test accuracy among five schemes, namely Local Training, FedAvg, APFL, SCEI-Async, and \emph{SCEI}, is presented in Fig.~\ref{fig:acc_sota}. \emph{SCEI} consistently outperforms the other schemes. Specifically, when training CNN on CIFAR-10, \emph{SCEI} achieves higher average local test accuracy than Local Training, APFL, FedAvg, SCEI-Async by $3\%$, $6\%$, $25\%$, and $3\%$, respectively. Notably, when training CNN on UCI and REALWORLD datasets, the improvement of \emph{SCEI} in average local test accuracy is still noticeable. This improvement is more prominent due to the smaller number of classes ($6$ and $8$, respectively) in these datasets, which reduces the classification difficulty. By contrast, when training CNN on the more challenging IMAGENET dataset with $1000$ classes, \emph{SCEI} exhibits substantial improvement in model accuracy compared to state-of-the-art schemes. This verifies the effectiveness of \emph{SCEI} on complex datasets. By comparing with CNN, when training a more complex model RESNET on CIFAR-10, \emph{SCEI} achieves $10\%$ higher model accuracy than APFL, FedAvg, and Local Training, because a more complex model allows \emph{SCEI} to learn more features from each node. This also reveals that \emph{SCEI} has a greater potential in utilizing the complex neural network. Besides, \emph{SCEI} outperforms Local Training in average local test accuracy, because the model trained on each node learns important features from the local models of other nodes. FedAvg always achieves the lowest average local test accuracy due to the inadequate adaptation of the global model to non-iid data.

Compared to Local Training and APFL, the improvement of \emph{SCEI} in average local test accuracy is not as significant when training models on the MNIST dataset, because the MNIST dataset is the simplest among all the datasets. As depicted in Fig.~\ref{fig:acc_sota}, Local Training, serving as a benchmark, produces higher average local test accuracy (about $96\%$) when training CNN on MNIST than the other datasets. Besides, MLP exhibits a lower average local test accuracy than CNN due to its limited expressive capacity. Therefore, \emph{SCEI} consistently demonstrates a stable performance improvement over state-of-the-art schemes despite the simplicity of the dataset.

In SCEI-Async, the asynchronous aggregation strategy leads to the calculation of the optimal $\alpha$ based on outdated model accuracies of nodes. Since the decision on the optimal $\alpha$ can be influenced by outdated model accuracies from slower nodes, SCEI-Async consistently achieves lower model accuracy than SCEI, as shown in Fig.~\ref{fig:acc_sota}.

\begin{figure*}
  \begin{center}
  \includegraphics[width=0.8\textwidth]{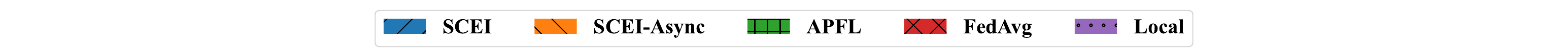}
    \subfigure[CNN on CIFAR-10]{
      \includegraphics[width=0.24\textwidth]{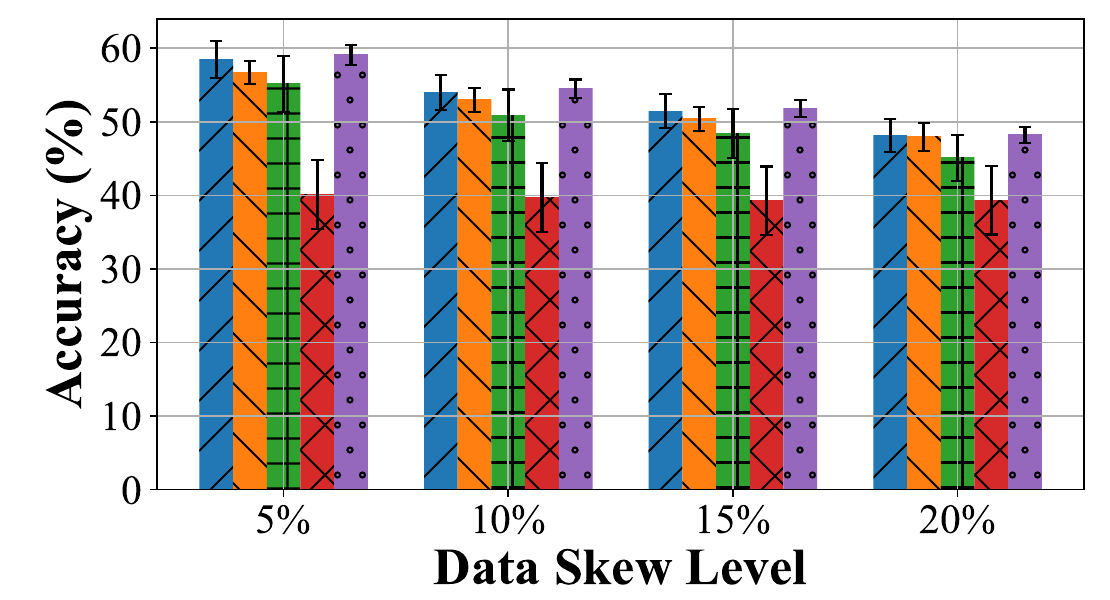}}
    \subfigure[RESNET on CIFAR-10]{
      \includegraphics[width=0.24\textwidth]{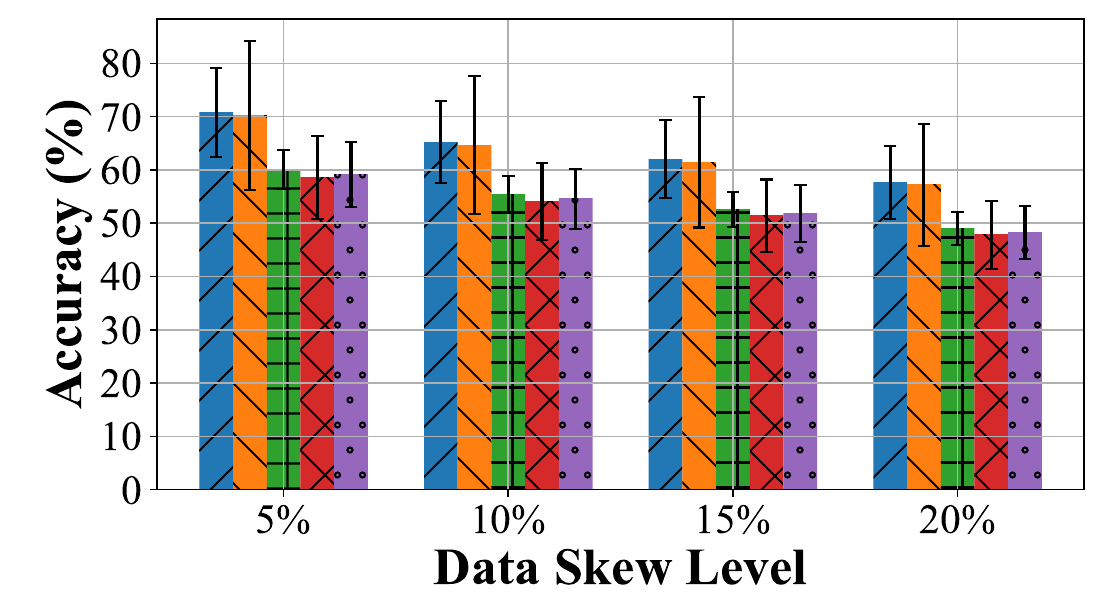}}
    \subfigure[CNN on MNIST]{
      \includegraphics[width=0.24\textwidth]{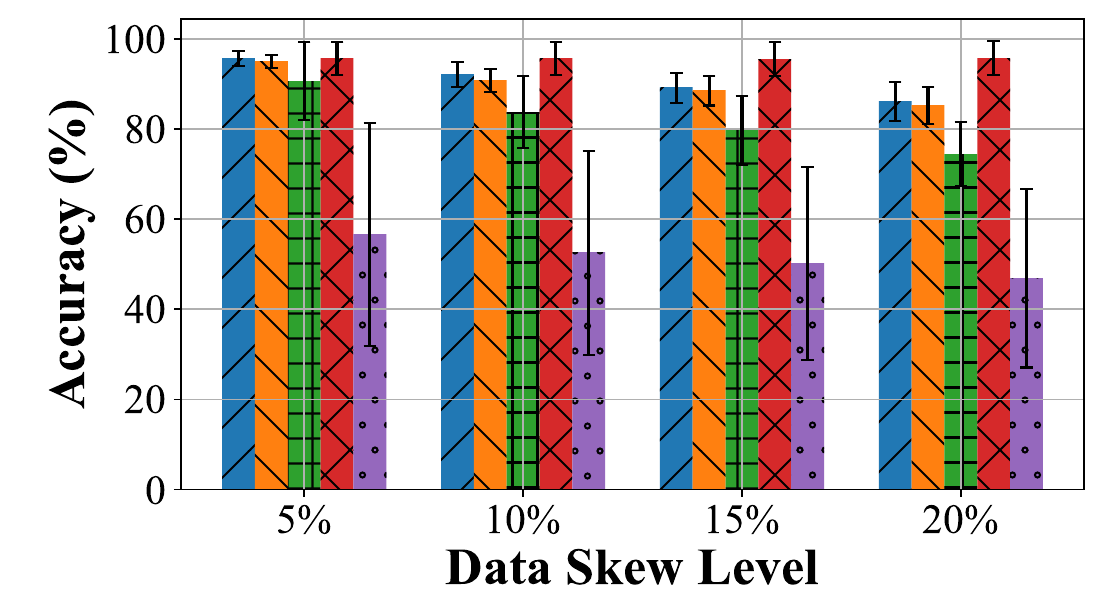}}
    \subfigure[MLP on MNIST]{
      \includegraphics[width=0.24\textwidth]{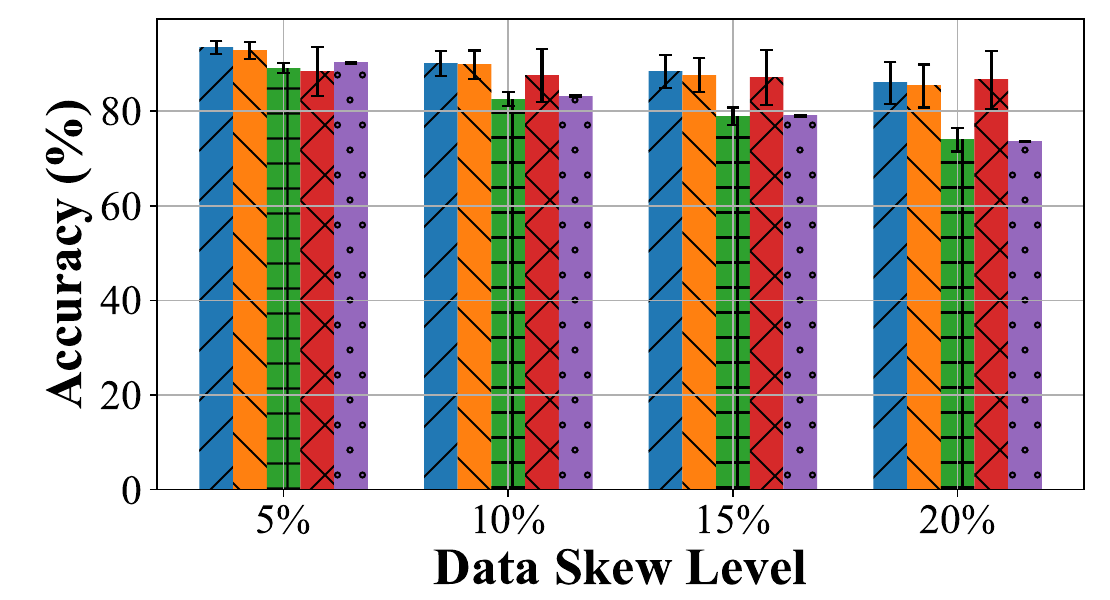}}
    \subfigure[CNN on CIFAR-100]{
      \includegraphics[width=0.24\textwidth]{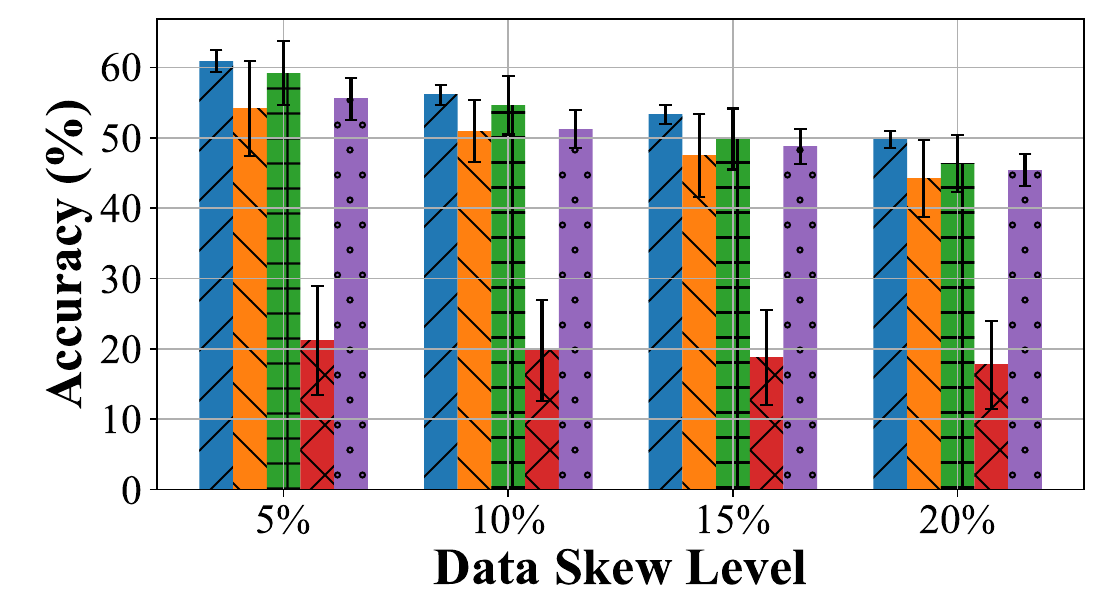}}
    \subfigure[CNN on UCI]{
      \includegraphics[width=0.24\textwidth]{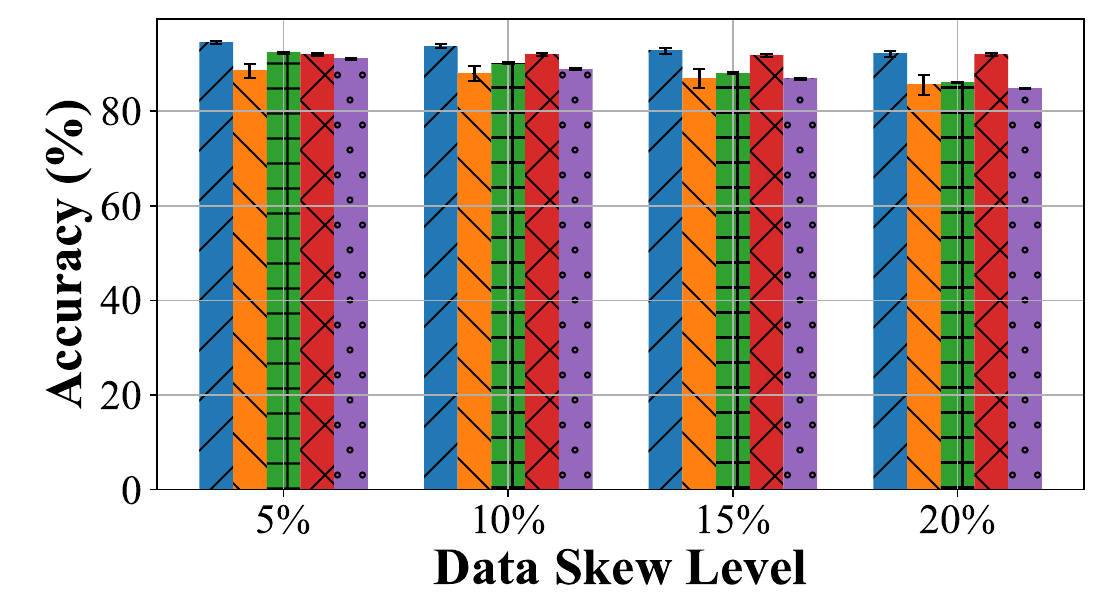}}
    \subfigure[CNN on REALWORLD]{
      \includegraphics[width=0.24\textwidth]{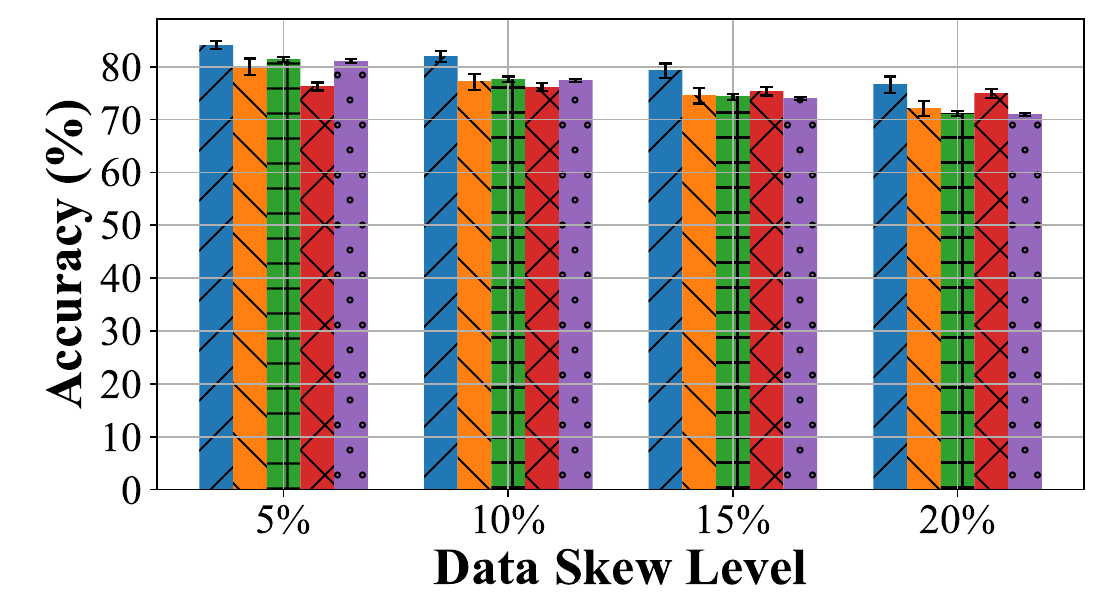}}
    \subfigure[CNN on IMAGENET]{
      \includegraphics[width=0.24\textwidth]{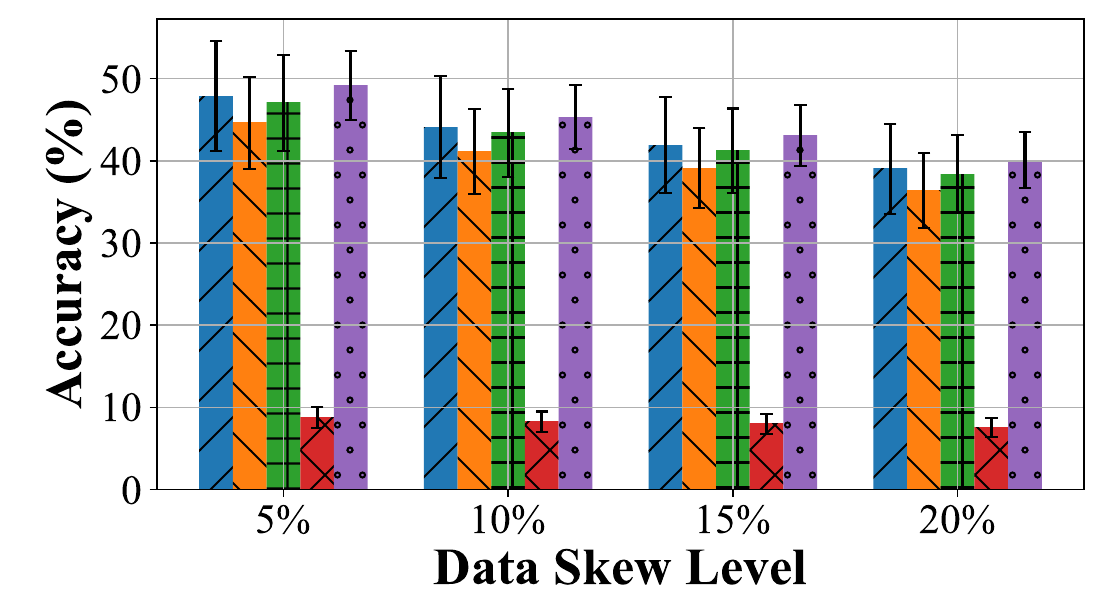}}
  \end{center}
  \caption{Comparison of average local test accuracy between SCEI and other schemes under varying levels of data skewness.}
  \label{fig:acc_skew}
\end{figure*}

In general, as $\alpha$ approaches $1$, the model converges faster and achieves higher accuracy. However, as $\alpha$ increases, the model's generalization capability declines, causing the personalized model to overfit the local training data. Such personalized models struggle to handle new data that does not fit the local data distribution. To investigate this, experiments are conducted by introducing skewed data ranging from $5\%$ to $20\%$ to each node. As depicted in Fig.~\ref{fig:acc_skew}, the accuracy of models trained on locally skewed data (at levels of $5\%$, $10\%$, $15\%$, and $20\%$) is compared. With an increase in skewed data, the model accuracy for all schemes decreases to varying extents. Specifically, FedAvg demonstrates the slowest rate of performance degradation since it is built for an ideal global model that can handle skewed data effectively. However, \emph{SCEI} consistently outperforms other state-of-the-art schemes across all data-skew scenarios due to its federated optimization approach.

Two further findings are worth noticing.
Firstly, when there are $5\%$ skewed data, \emph{SCEI} demonstrates better model accuracy than FedAvg regardless of the model and dataset. This highlights the suitability of \emph{SCEI} for scenarios with low data skew, which are prevalent in real-world scenarios. For example, in a smart hospital, about $5\%$ of patients may require referrals to other facilities due to complex conditions. Similarly, on a smart farm, around $5\%$ of crops may be replaced with new varieties every year. Moreover, if the proportion of skewed data exceeds $5\%$, one possible solution is to adjust $\alpha$ to zero and switches \emph{SCEI} to FedAvg mode to obtain a more generalized global model.
Secondly, when training the CNN model on complex datasets, such as CIFAR-100, UCI, REALWORLD, and IMAGENET, \emph{SCEI} consistently outperforms other schemes regardless of the data skew level. This indicates that \emph{SCEI} is the most effective solution for achieving a balance between personalized and federated models in scenarios involving complex datasets with skewed data.

\begin{framed}
\noindent Result 1: 
\emph{SCEI} outperforms state-of-the-art schemes in terms of model accuracy, particularly when confronted with complex datasets. Besides, \emph{SCEI} is capable of dealing with data skew, a common challenge faced by personalized model training approaches.
\end{framed}

\subsubsection{RQ2. Scalability}

\begin{figure*}
  \begin{center}
  \includegraphics[width=0.8\textwidth]{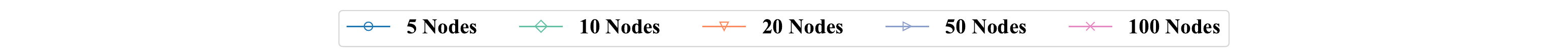}
    \subfigure[CNN on CIFAR-10]{
      \includegraphics[width=0.24\textwidth]{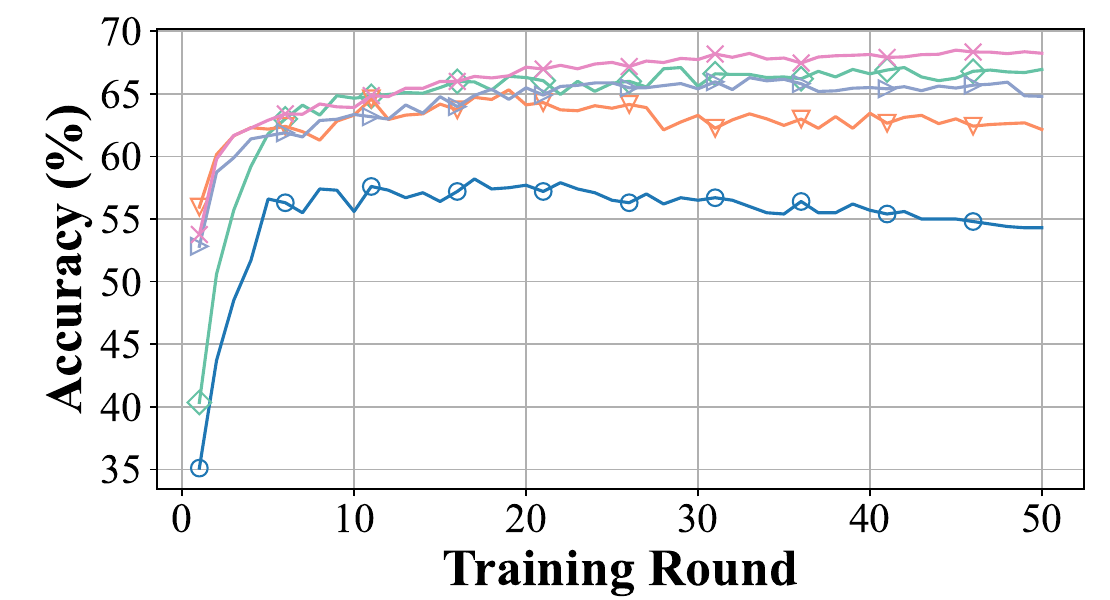}}
    \subfigure[RESNET on CIFAR-10]{
      \includegraphics[width=0.24\textwidth]{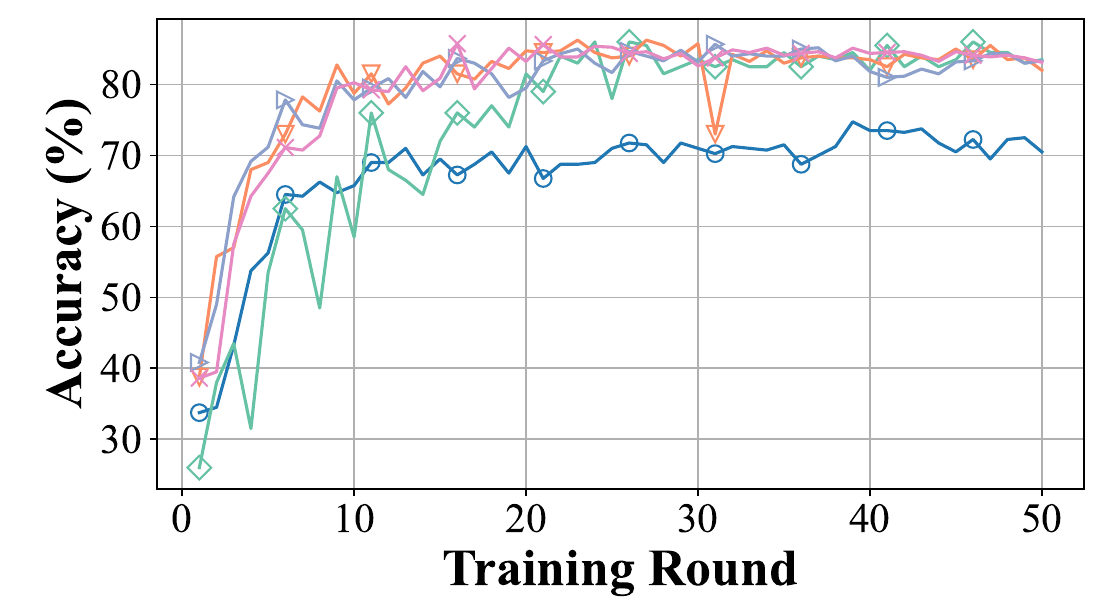}}
    \subfigure[CNN on MNIST]{
      \includegraphics[width=0.24\textwidth]{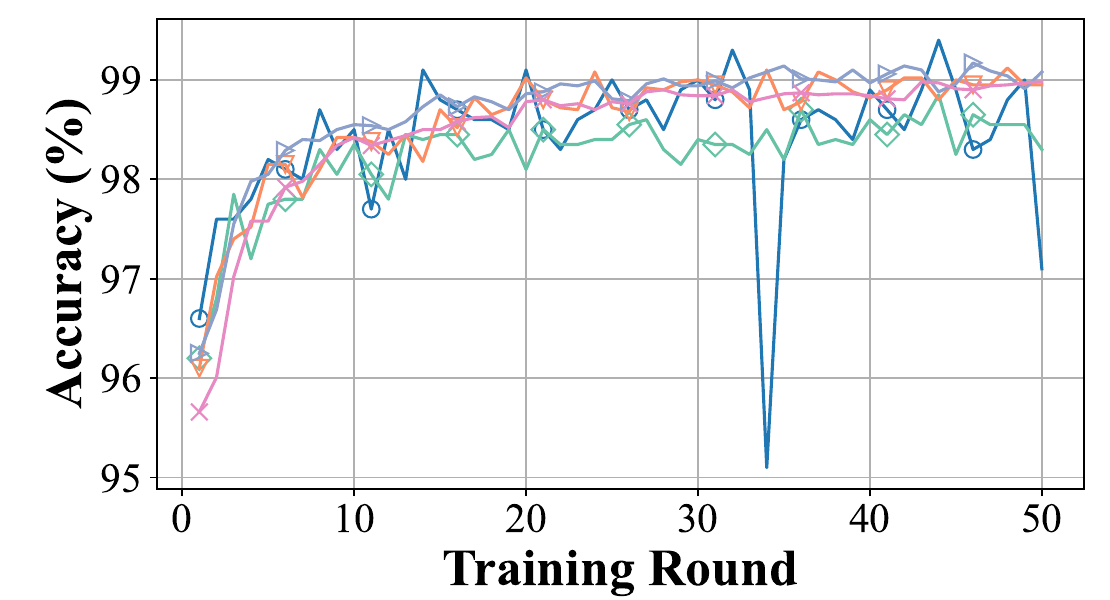}}
    \subfigure[MLP on MNIST]{
      \includegraphics[width=0.24\textwidth]{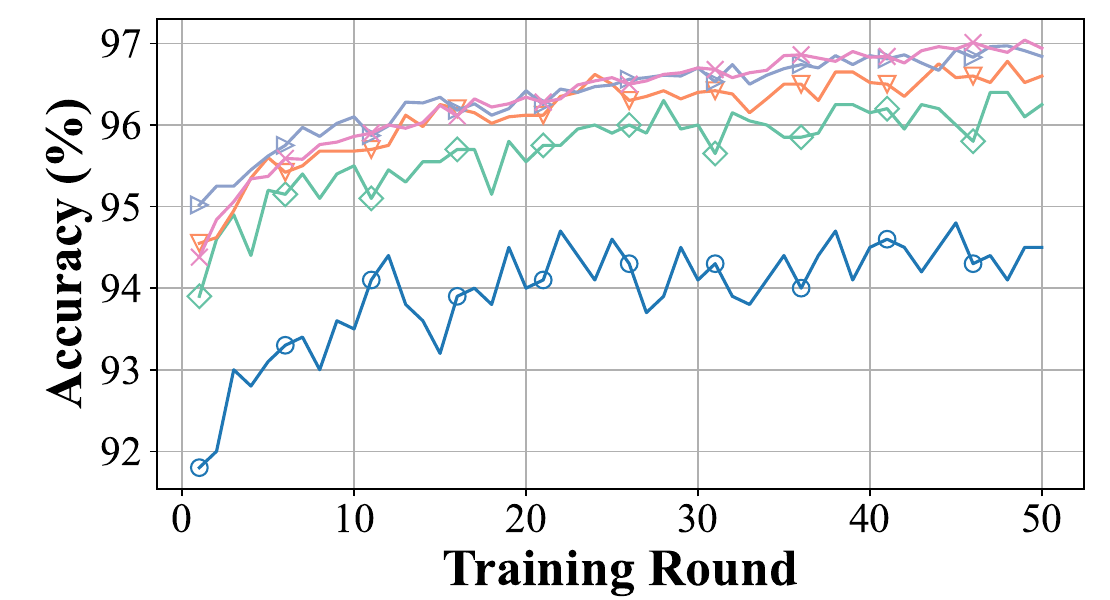}}
    \subfigure[CNN on CIFAR-100]{
      \includegraphics[width=0.24\textwidth]{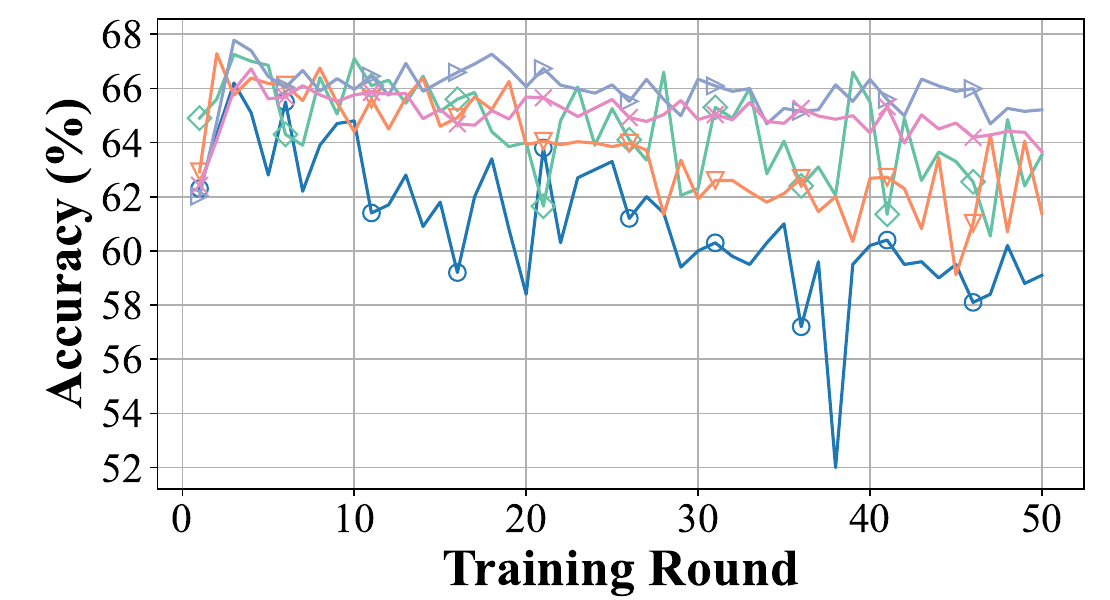}}
    \subfigure[CNN on UCI]{
      \includegraphics[width=0.24\textwidth]{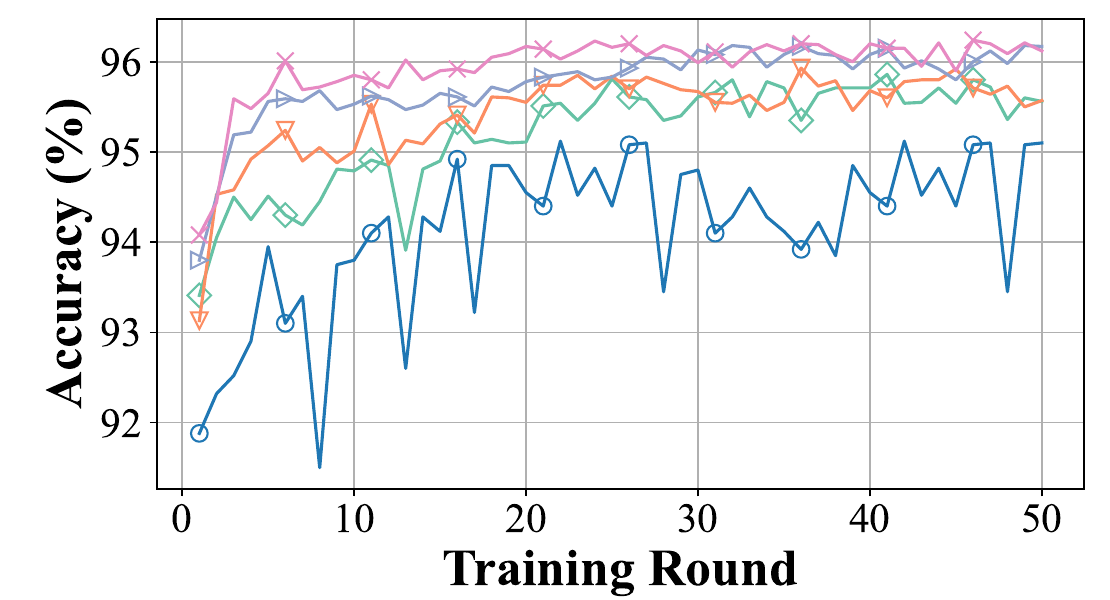}}
    \subfigure[CNN on REALWORLD]{
      \includegraphics[width=0.24\textwidth]{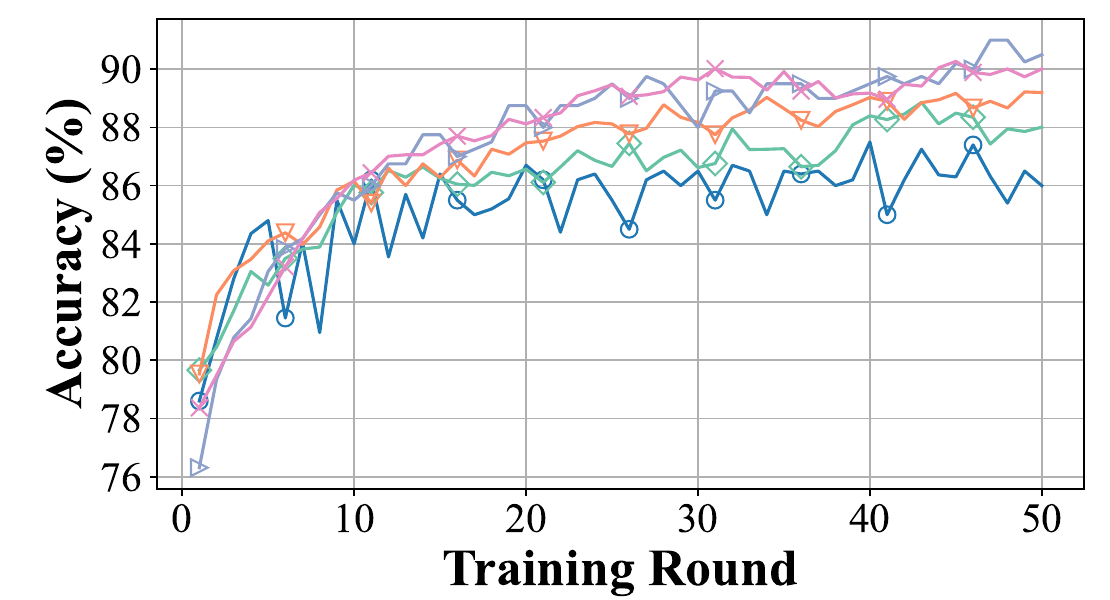}}
    \subfigure[CNN on IMAGENET]{
      \includegraphics[width=0.24\textwidth]{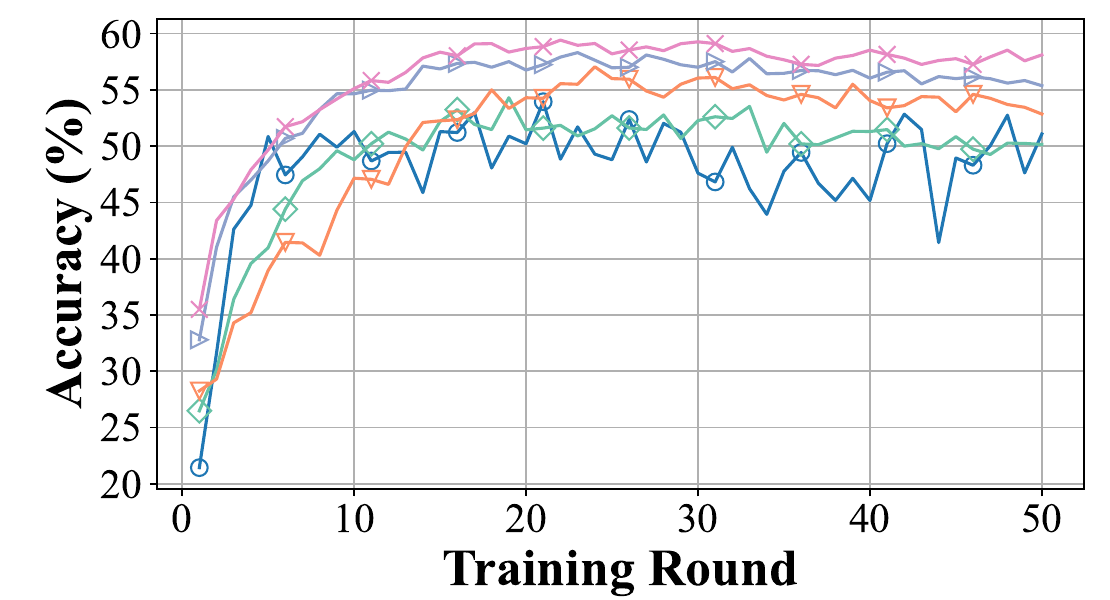}}
  \end{center}
  \caption{Comparison of average local test accuracy for SCEI across different numbers of nodes.}
  \label{fig:acc_nodes}
\end{figure*}

\begin{figure*}
  \begin{center}
  \includegraphics[width=0.8\textwidth]{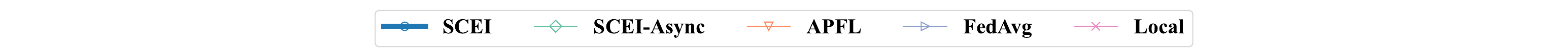}
    \subfigure[CNN on CIFAR-10]{
      \includegraphics[width=0.24\textwidth]{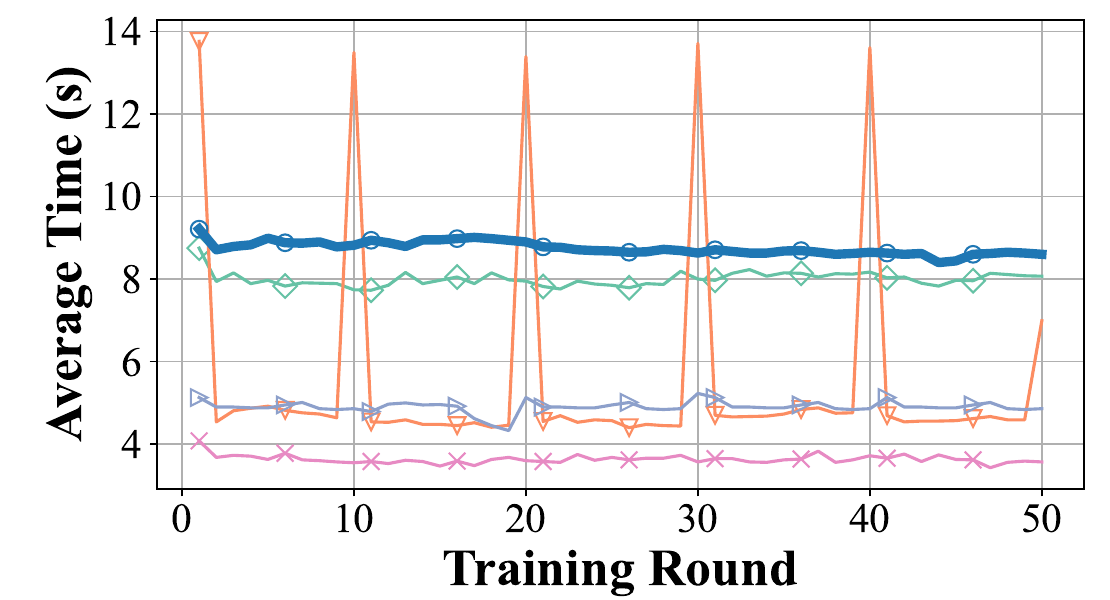}}
    \subfigure[RESNET on CIFAR-10]{
      \includegraphics[width=0.24\textwidth]{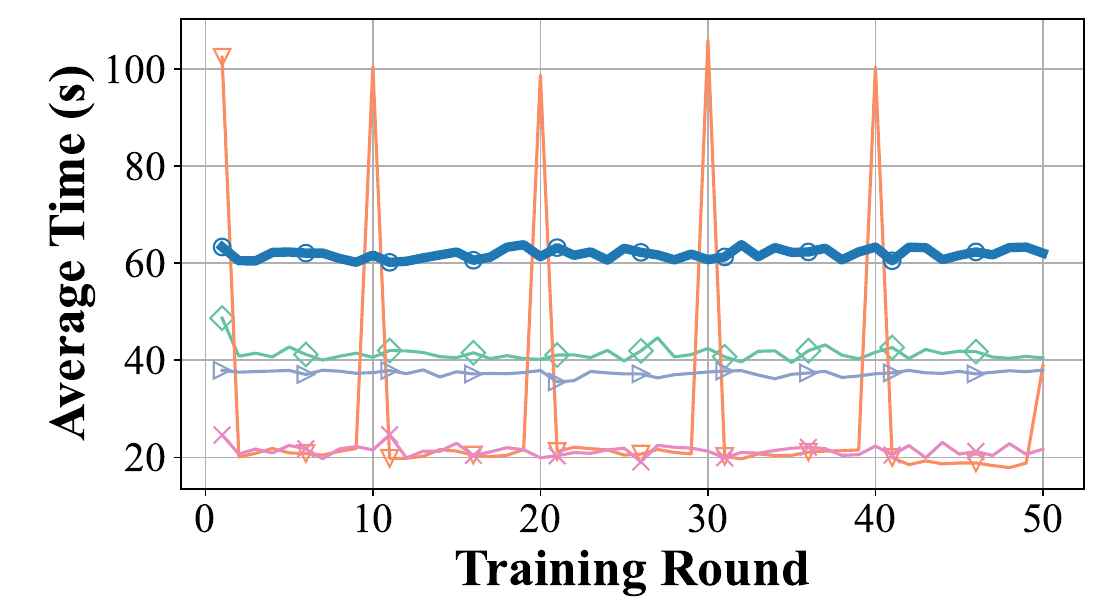}}
    \subfigure[CNN on MNIST]{
      \includegraphics[width=0.24\textwidth]{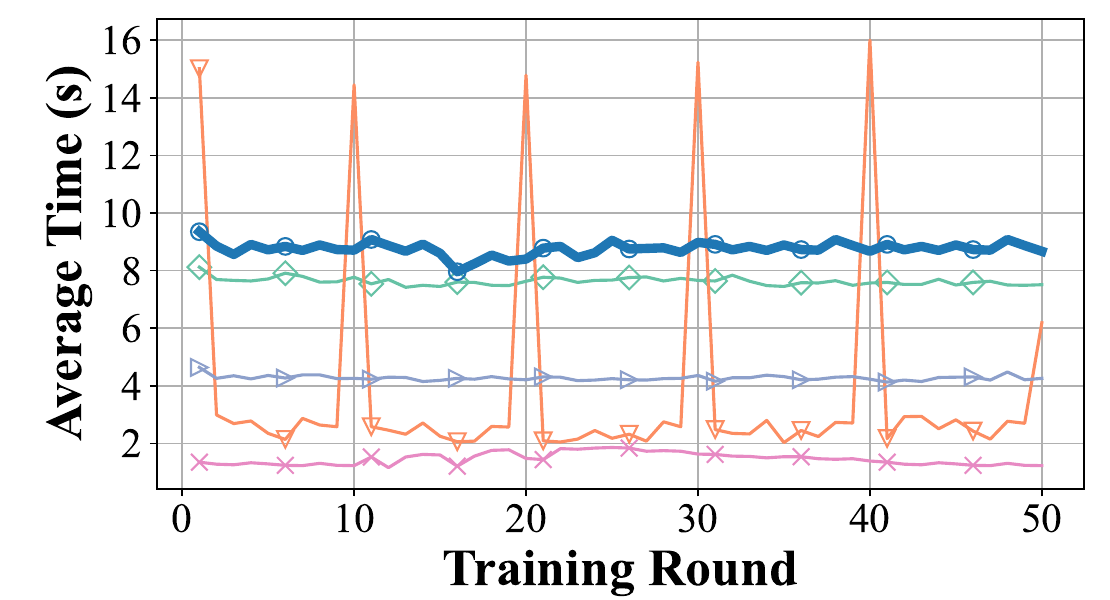}}
    \subfigure[MLP on MNIST]{
      \includegraphics[width=0.24\textwidth]{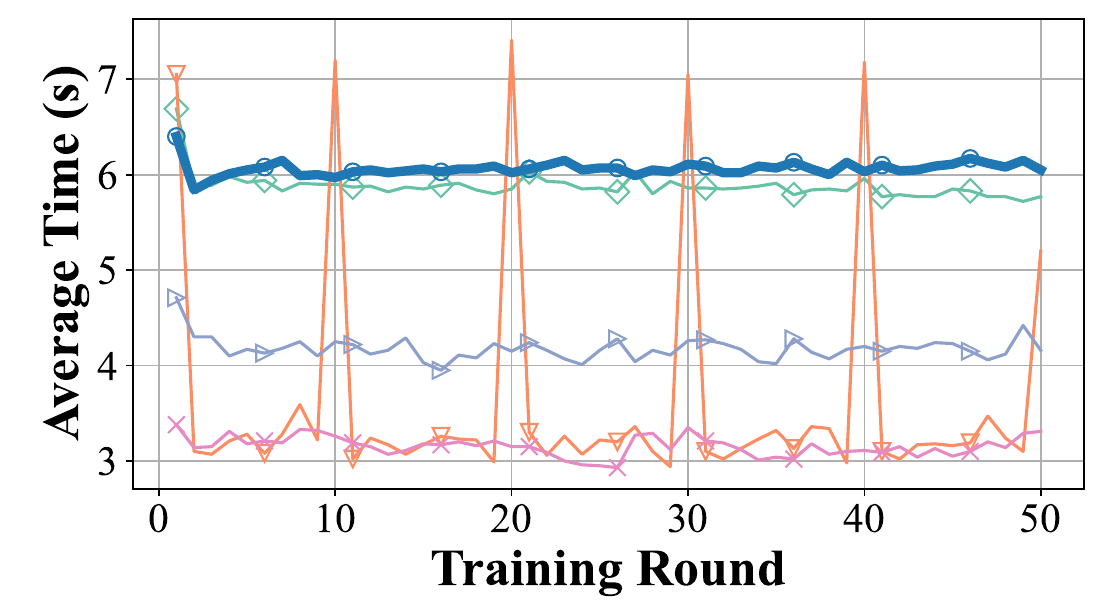}}
    \subfigure[CNN on CIFAR-100]{
      \includegraphics[width=0.24\textwidth]{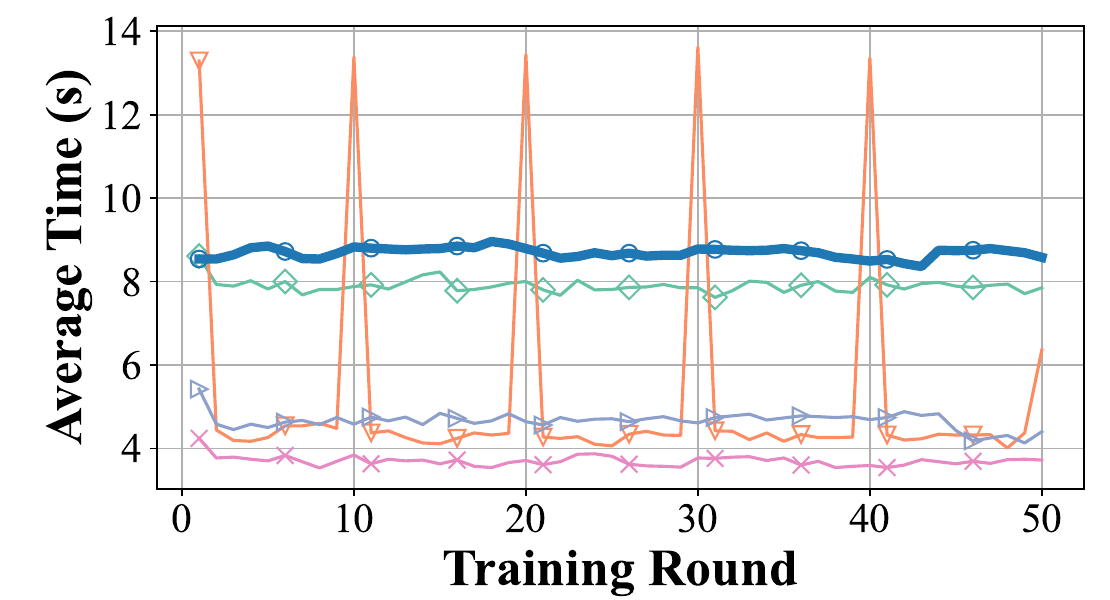}}
    \subfigure[CNN on UCI]{
      \includegraphics[width=0.24\textwidth]{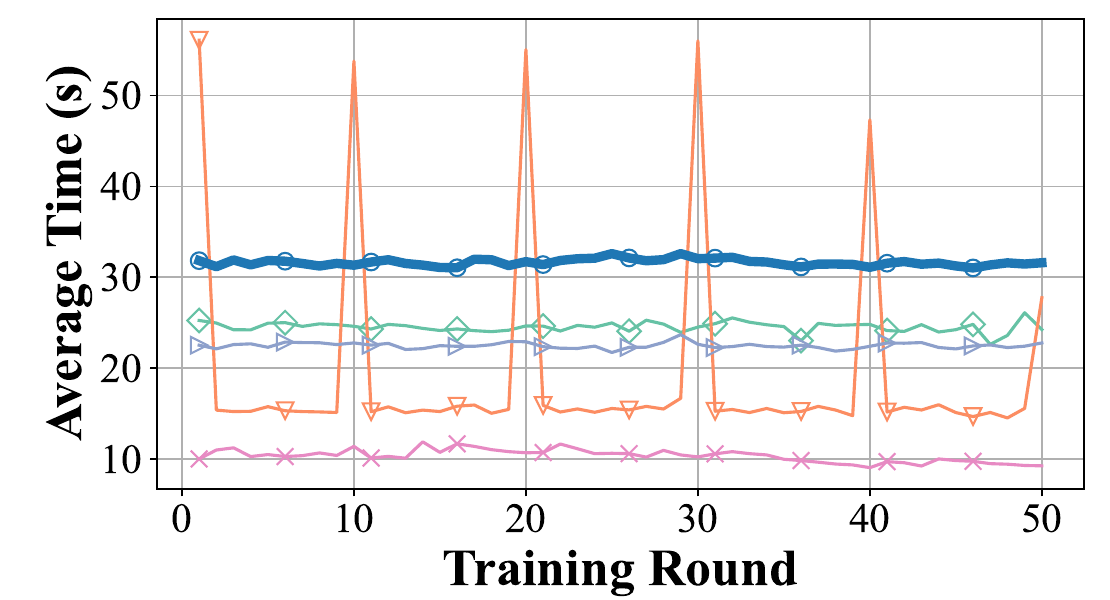}}
    \subfigure[CNN on REALWORLD]{
      \includegraphics[width=0.24\textwidth]{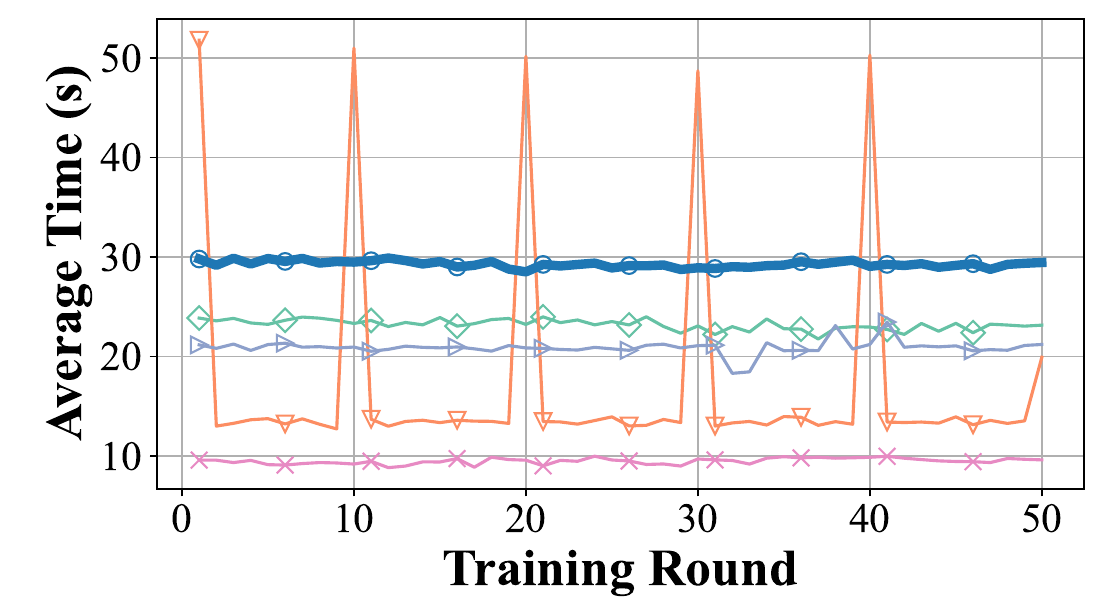}}
    \subfigure[CNN on IMAGENET]{
      \includegraphics[width=0.24\textwidth]{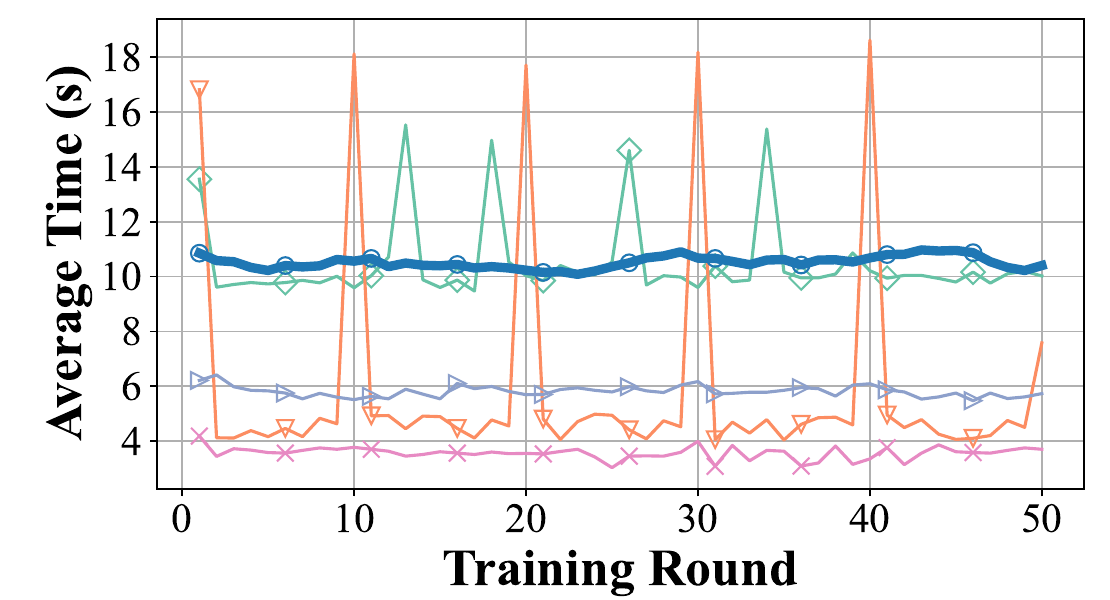}}
  \end{center}
  \caption{Comparison of average overall time cost per training round for SCEI, FedAvg, APFL, and Local Training.}
  \label{fig:time_overall}
\end{figure*}

\begin{figure*}
  \begin{center}
  \includegraphics[width=0.8\textwidth]{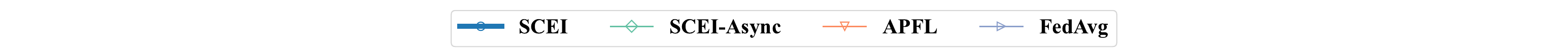}
    \subfigure[CNN on CIFAR-10]{
      \includegraphics[width=0.24\textwidth]{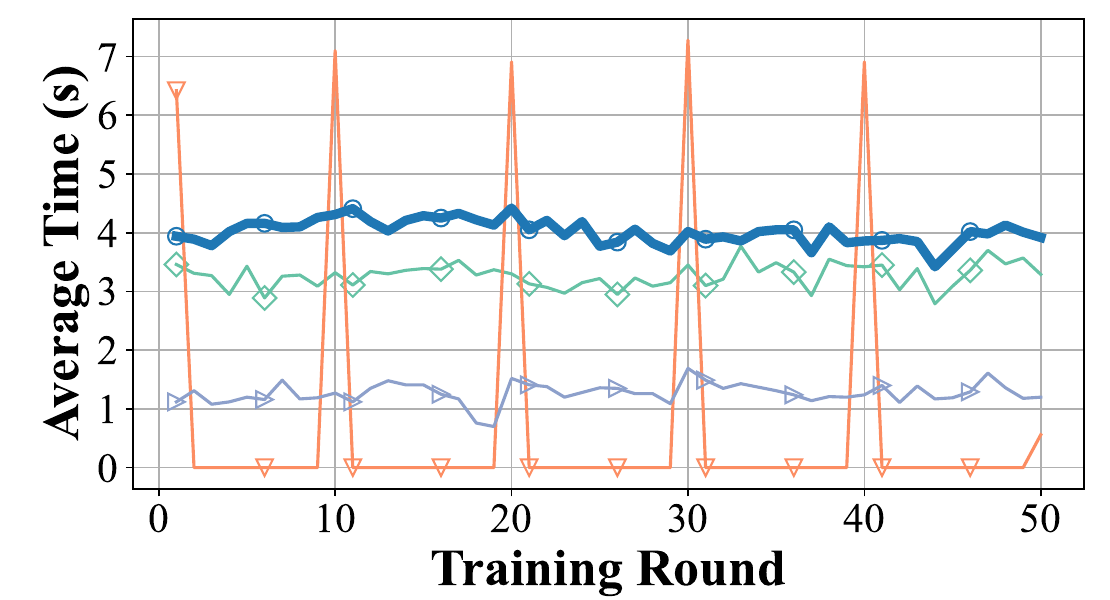}}
    \subfigure[RESNET on CIFAR-10]{
      \includegraphics[width=0.24\textwidth]{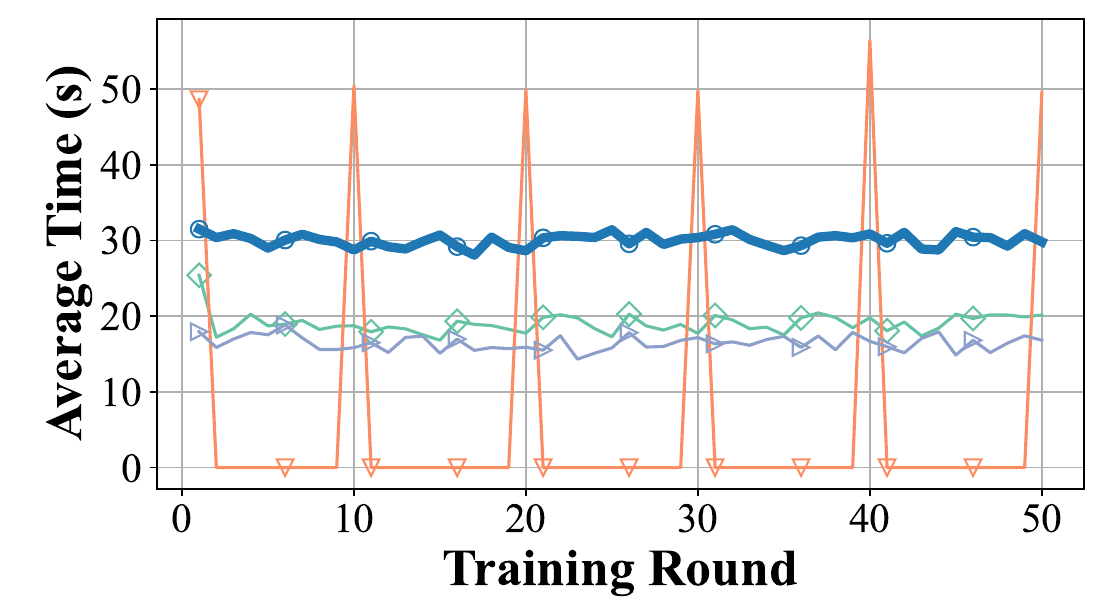}}
    \subfigure[CNN on MNIST]{
      \includegraphics[width=0.24\textwidth]{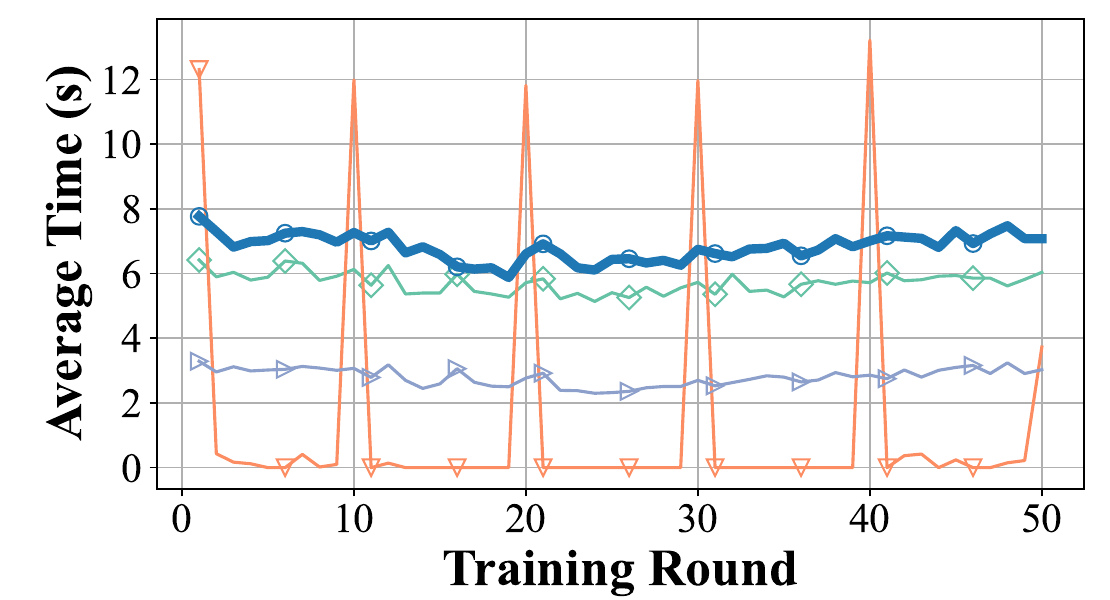}}
    \subfigure[MLP on MNIST]{
      \includegraphics[width=0.24\textwidth]{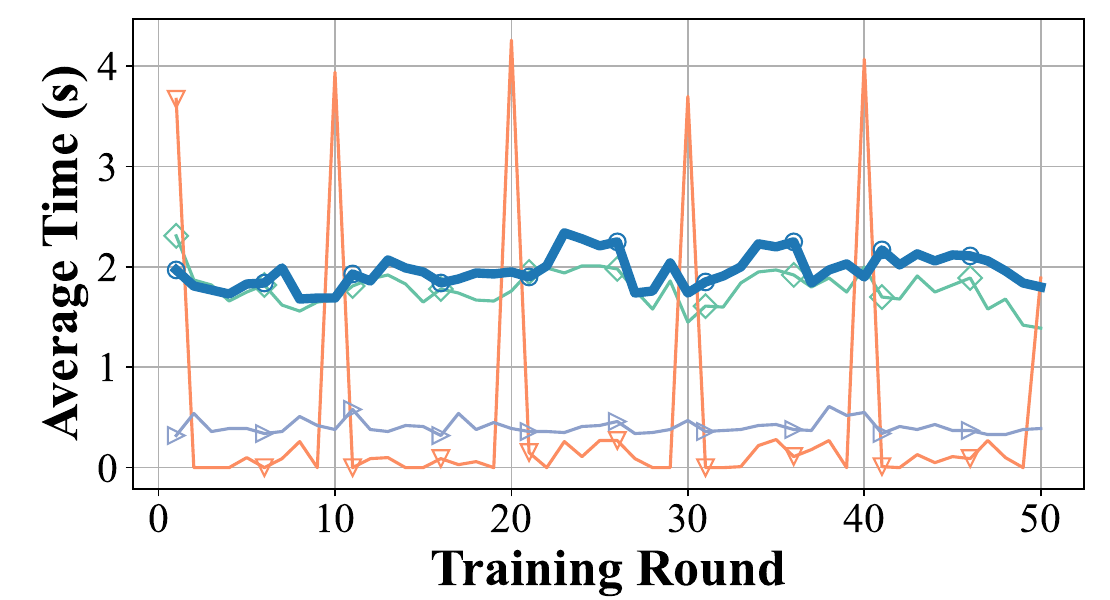}}
    \subfigure[CNN on CIFAR-100]{
      \includegraphics[width=0.24\textwidth]{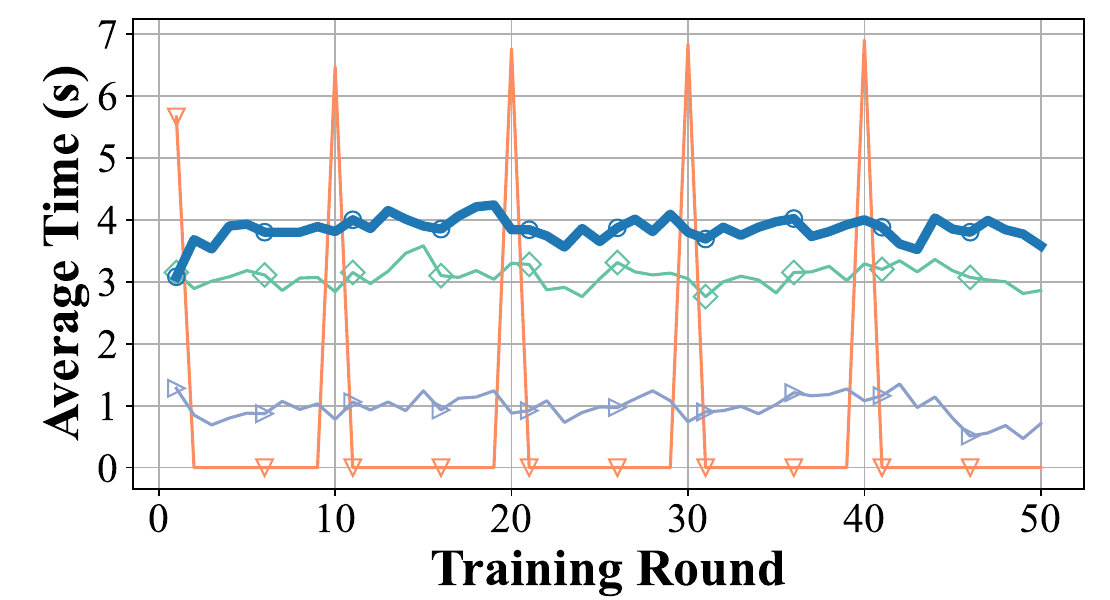}}
    \subfigure[CNN on UCI]{
      \includegraphics[width=0.24\textwidth]{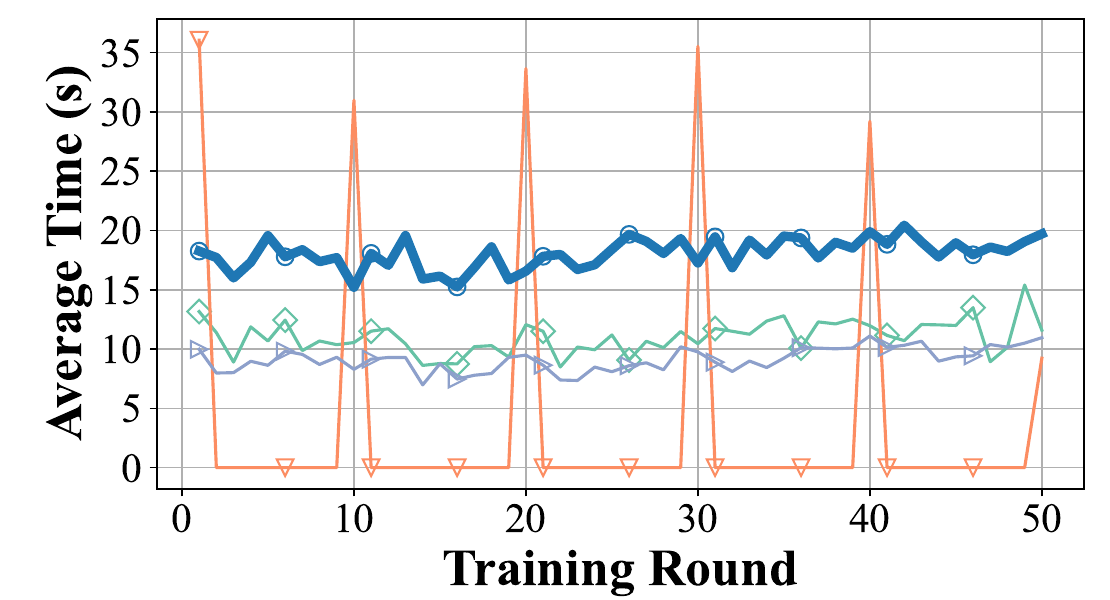}}
    \subfigure[CNN on REALWORLD]{
      \includegraphics[width=0.24\textwidth]{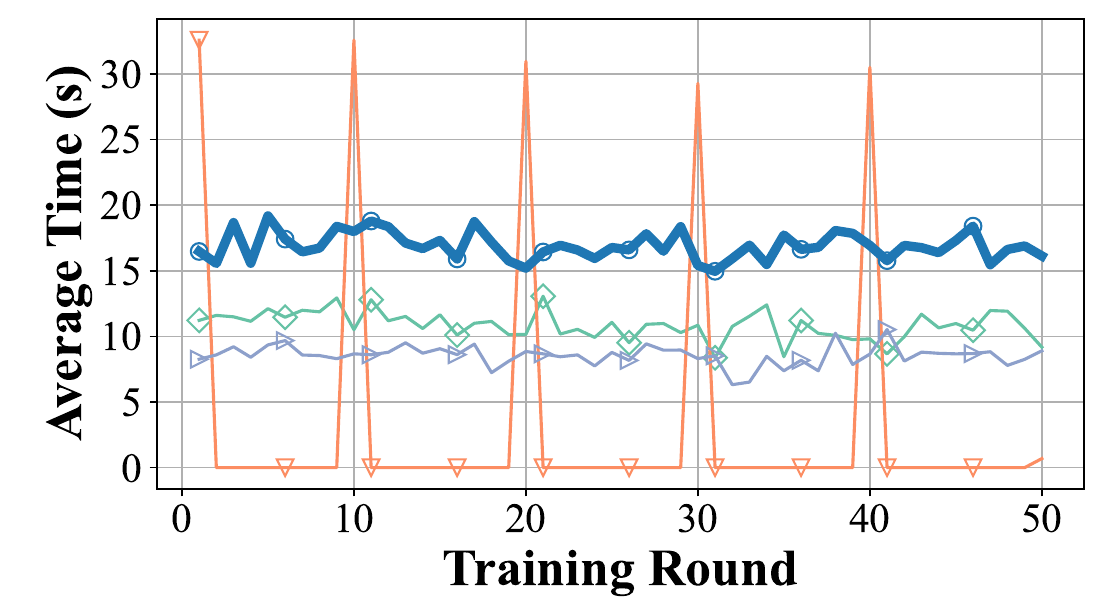}}
    \subfigure[CNN on IMAGENET]{
      \includegraphics[width=0.24\textwidth]{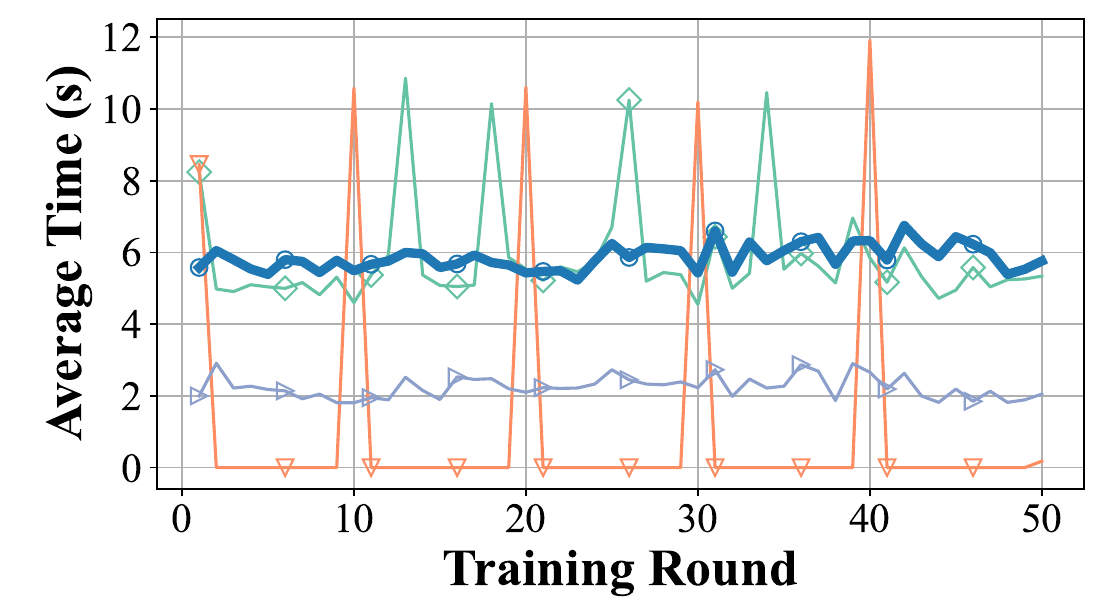}}
  \end{center}
  \caption{Comparison of average communication time cost per training round for SCEI, FedAvg, and APFL.}
  \label{fig:time_commu}
\end{figure*}

There is a balance between model accuracy and costs. As shown in Fig.~\ref{fig:acc_nodes}, \emph{SCEI} achieves higher average local test accuracy with an increased number of nodes (as well as increased computing and communication costs). This improvement is due to the inclusion of more nodes with diverse data, which improves the accuracy of personalized local models in non-iid scenarios. However, the rate of improvement in average local test accuracy diminishes when the number of nodes surpasses $20$, the ideal node number. For example, when training CNN on CIFAR-10, increasing the number of nodes to $10$ increases the accuracy by $10\%$, whereas increasing it to $100$ only increases the accuracy by $5\%$. This is because the CNN model reaches a saturation point in terms of learning new features from federated optimization after learning sufficient features from a certain number of nodes. Notably, the ideal node number $20$ is application specific. For instance, as the IMAGENET dataset encompasses the largest number of classes and features to learn, \emph{SCEI} consistently demonstrates enhanced performance as the number of nodes increases. Therefore, the ideal node number may increase if the model, dataset, and application scenario become more complicated.

When training CNN on MNIST with an expanding number of nodes ranging from $5$ to $100$, the increase in average local test accuracy is not obvious. The reason for this is that the CNN model is advanced enough to fit the MNIST dataset well and achieve high average local test accuracy, even with a limited amount of training data. By contrast, when training MLP on MNIST, the increase in the average local test accuracy is more noticeable (about $3\%$) as the number of nodes increases from $5$ to $100$. This disparity arises due to the simpler architecture of the MLP model, which requires a larger amount of data to achieve higher accuracy. In fact, when comparing training the MLP model with $100$ nodes to training the CNN model with $5$ nodes on the MNIST dataset, CNN easily attains superior model accuracy.

\begin{framed}
\noindent Result 2: 
\emph{SCEI} quickly converges to an acceptable average local test accuracy and shows the potential for a better personalized local model with additional local data.
\end{framed}

\subsubsection{RQ3. Computation and Communication Overhead}

As shown in Fig.~\ref{fig:time_overall}, Local Training always exhibits the lowest overall time cost in each training round due to no communication involved. Therefore, Local Training serves as a benchmark for the other three schemes. However, when training CNN on the UCI and REALWORLD datasets, the overall time cost of Local Training is significantly higher (around $10$ seconds) compared to other datasets. This discrepancy can be attributed to the increased complexity of these datasets and CNN models. Generally, the more perceptrons there are in the model, the greater the number of weights to be tuned. The back-propagation algorithm employed for weight adjustment is performed for every training example and is repeated over numerous local training rounds. Consequently, increasing the number of perceptrons, training samples, and local training rounds leads to more weight adjustments, resulting in greater training time costs. Additionally, different neural network architectures contribute to varying computational workloads. For instance, in the case of CNN, the size of filters employed significantly impacts the computational workload. A $10\times10$ filter entails adjusting $100$ parameters, while a $3\times3$ filter requires only $9$ parameter adjustments. Consider an image with dimensions of $150\times150$ and a stride size of $1$. With a $10\times10$ filter, the filter would be applied $140\times140=19,600$ times, thus necessitating a total adjustment of $19,600\times100=1,960,000$ parameters. On the other hand, a $3\times3$ filter would be applied $147\times147=21,609$ times, resulting in a total adjustment of $21,609\times9=194,481$ parameters. However, for MLP, the concept of filters does not apply, and ResNet incorporates shortcuts to bypass the computation of weights in certain layers.

Despite the variations among the models, when focused on a particular model and dataset, FedAvg usually incurs a higher average overall time cost than Local Training due to the additional steps of model parameter transmission and global model aggregation. Similarly, APFL requires more time than Local Training due to its two-phase training approach and periodic communication every $10$ rounds. \emph{SCEI} maintains an acceptable average overall time cost of less than $30$ seconds in most cases, making it suitable for scenarios like smart hospitals and smart farms. When training CNN on the UCI and REALWORLD datasets, \emph{SCEI} takes about $10$ seconds longer than FedAvg per round, which is reasonable considering the improved model accuracy achieved on skewed data. By using an asynchronous aggregation strategy, SCEI-Async reduces the waiting time for other nodes, resulting in a lower average overall time cost than \emph{SCEI}, at the expense of model accuracy.

As shown in Fig.~\ref{fig:time_commu}, both \emph{SCEI} and SCEI-async exhibit an average communication time cost of no more than $20$ seconds per round, which is higher than that of FedAvg. This increase is due to the additional consensus procedure imposed by the blockchain. However, as the complexity of the model grows, such as when training the RESNET model on the CIFAR-10 dataset, the average communication time cost of \emph{SCEI} per round grows to roughly $30$ seconds. This is attributed to the heavier computational workload involved in the training process, which slows down the consensus, model transmission, and encryption/decryption steps. One way to improve efficiency is to reduce the waiting time for other nodes by adopting an asynchronous aggregation strategy, like SCEI-async. When dealing with more complex datasets and models, SCEI-async shows a greater improvement in terms of time efficiency (reducing by around $10$ seconds per training round). However, this improvement comes at the expense of model accuracy, as SCEI-async achieves around $4\%$ lower model accuracy than SCEI when training RESNET on CIFAR-10, as shown in Fig.~\ref{fig:acc_sota}. Reducing the committee size is another approach to decrease the communication time cost, although this compromises the security of the blockchain to some extent. In addition, since APFL requires model aggregations every $10$ training rounds, the fastest node must wait for up to $9$ rounds for the slowest node, resulting in longer communication time than \emph{SCEI}.

Besides, for SCEI, there is no trade-off between training time and communication time. Generally, a more complex model will cause a higher training time and higher communication time, as more parameters are required to be transmitted. This insight can be observed by comparing the results in Fig.~\ref{fig:time_overall} and Fig.~\ref{fig:time_commu}.

\begin{framed}
\noindent Result 3: In comparison to state-of-the-art schemes, \emph{SCEI} exhibits a reasonable training overhead primarily due to its communication requirements within the secure blockchain infrastructure. However, this overhead remains relatively consistent and is lower than that observed in recent research on personalized models.
\end{framed}

\subsection{Discussion}
The proposed scheme is a general framework that works effectively in cross-silo FL scenarios with reliable communication~\cite{mcmahan2021advances, li2020privacy}. To the best knowledge, the proposed framework is promising in terms of model accuracy, scalability, credibility, and security. In the experiments, as samples of each node are retrieved from four different classes and each class represents $25\%$ of the total, data samples with a skew of over $20\%$ are no longer considered skewed. Besides, the experiment results highlight that \emph{SCEI} outperforms state-of-the-art schemes when dealing with complex models and datasets, even when a moderate level of data skew is present. However, due to resource and hardware limitations, experiments were not conducted in a real-world mobile IoT environment with realistic and ad-hoc communication time for nodes~\cite{zhang2020democratically}.

\section{Conclusion}

To address the challenges posed by non-iid and skewed data, as well as address trust and security concerns in distributed learning schemes in IoT systems, a smart contract-driven edge intelligence framework (SCEI) is proposed. SCEI introduces a novel personalized model training scheme by leveraging smart contracts to facilitate federated optimization across participating learning nodes. An open-sourced prototype of SCEI has been implemented, and experimental evaluations have been conducted using various learning models and datasets.   The results demonstrate that SCEI outperforms state-of-the-art personalized learning schemes when handling skewed data. Future work will involve further assessments of the proposed framework using diverse models, datasets, and real-world multi-modal sensor data.

\section*{Acknowledgment}

Supported by the Taishan Scholars Program No. TSQN202211214 and Shandong Excellent Young Scientists Fund Program (Overseas) No.2023HWYQ-113. 

Many thanks to Wanping Bai for her valuable time and contributions in enhancing the quality of this paper.

\bibliography{references}{}

\begin{thebibliography}{10}
\providecommand{\url}[1]{#1}
\csname url@samestyle\endcsname
\providecommand{\newblock}{\relax}
\providecommand{\bibinfo}[2]{#2}
\providecommand{\BIBentrySTDinterwordspacing}{\spaceskip=0pt\relax}
\providecommand{\BIBentryALTinterwordstretchfactor}{4}
\providecommand{\BIBentryALTinterwordspacing}{\spaceskip=\fontdimen2\font plus
\BIBentryALTinterwordstretchfactor\fontdimen3\font minus
  \fontdimen4\font\relax}
\providecommand{\BIBforeignlanguage}[2]{{%
\expandafter\ifx\csname l@#1\endcsname\relax
\typeout{** WARNING: IEEEtran.bst: No hyphenation pattern has been}%
\typeout{** loaded for the language `#1'. Using the pattern for}%
\typeout{** the default language instead.}%
\else
\language=\csname l@#1\endcsname
\fi
#2}}
\providecommand{\BIBdecl}{\relax}
\BIBdecl

\bibitem{srivastava2019deep}
A.~Srivastava, S.~Sengupta, S.-J. Kang, K.~Kant, M.~Khan, S.~A. Ali, S.~R.
  Moore, B.~C. Amadi, P.~Kelly, S.~Syed \emph{et~al.}, ``Deep learning for
  detecting diseases in gastrointestinal biopsy images,'' in \emph{2019 Systems
  and Information Engineering Design Symposium (SIEDS)}.\hskip 1em plus 0.5em
  minus 0.4em\relax IEEE, 2019, pp. 1--4.

\bibitem{boursianis2020internet}
A.~D. Boursianis, M.~S. Papadopoulou, P.~Diamantoulakis, A.~Liopa-Tsakalidi,
  P.~Barouchas, G.~Salahas, G.~Karagiannidis, S.~Wan, and S.~K. Goudos,
  ``Internet of things (iot) and agricultural unmanned aerial vehicles (uavs)
  in smart farming: a comprehensive review,'' \emph{Internet of Things}, p.
  100187, 2020.

\bibitem{mcmahan2017communication}
B.~McMahan, E.~Moore, D.~Ramage, S.~Hampson, and B.~A. y~Arcas,
  ``Communication-efficient learning of deep networks from decentralized
  data,'' in \emph{Artificial Intelligence and Statistics}.\hskip 1em plus
  0.5em minus 0.4em\relax PMLR, 2017, pp. 1273--1282.

\bibitem{yang2019federated}
Q.~Yang, Y.~Liu, T.~Chen, and Y.~Tong, ``Federated machine learning: Concept
  and applications,'' \emph{ACM Transactions on Intelligent Systems and
  Technology (TIST)}, vol.~10, no.~2, pp. 1--19, 2019.

\bibitem{wang2020optimizing}
H.~Wang, Z.~Kaplan, D.~Niu, and B.~Li, ``Optimizing federated learning on
  non-iid data with reinforcement learning,'' in \emph{IEEE INFOCOM 2020-IEEE
  Conference on Computer Communications}.\hskip 1em plus 0.5em minus
  0.4em\relax IEEE, 2020, pp. 1698--1707.

\bibitem{zhu2019deep}
L.~Zhu, Z.~Liu, and S.~Han, ``Deep leakage from gradients,'' \emph{Advances in
  Neural Information Processing Systems}, vol.~32, 2019.

\bibitem{deng2020adaptive}
Y.~Deng, M.~M. Kamani, and M.~Mahdavi, ``Adaptive personalized federated
  learning,'' \emph{arXiv preprint arXiv:2003.13461}, 2020.

\bibitem{wu2020personalized}
Q.~Wu, K.~He, and X.~Chen, ``Personalized federated learning for intelligent
  iot applications: A cloud-edge based framework,'' \emph{IEEE Open Journal of
  the Computer Society}, vol.~1, pp. 35--44, 2020.

\bibitem{lee2021opportunistic}
S.~Lee, X.~Zheng, J.~Hua, H.~Vikalo, and C.~Julien, ``Opportunistic federated
  learning: An exploration of egocentric collaboration for pervasive computing
  applications,'' in \emph{2021 IEEE International Conference on Pervasive
  Computing and Communications (PerCom)}.\hskip 1em plus 0.5em minus
  0.4em\relax IEEE, 2021, pp. 1--8.

\bibitem{xu2021asynchronous}
C.~Xu, Y.~Qu, Y.~Xiang, and L.~Gao, ``Asynchronous federated learning on
  heterogeneous devices: A survey,'' \emph{arXiv preprint arXiv:2109.04269},
  2021.

\bibitem{xu2022efficient}
C.~Xu, Y.~Qu, T.~H. Luan, P.~W. Eklund, Y.~Xiang, and L.~Gao, ``An efficient
  and reliable asynchronous federated learning scheme for smart public
  transportation,'' \emph{IEEE Transactions on Vehicular Technology}, vol.~72,
  no.~5, pp. 6584--6598, 2023.

\bibitem{warnat2021swarm}
S.~Warnat-Herresthal, H.~Schultze, K.~L. Shastry, S.~Manamohan, S.~Mukherjee,
  V.~Garg, R.~Sarveswara, K.~H{\"a}ndler, P.~Pickkers, N.~A. Aziz
  \emph{et~al.}, ``Swarm learning for decentralized and confidential clinical
  machine learning,'' \emph{Nature}, vol. 594, no. 7862, pp. 265--270, 2021.

\bibitem{xu2021light}
C.~Xu, Y.~Qu, T.~H. Luan, P.~W. Eklund, Y.~Xiang, and L.~Gao, ``A lightweight
  and attack-proof bidirectional blockchain paradigm for internet of things,''
  \emph{IEEE Internet of Things Journal}, vol.~9, no.~6, pp. 4371--4384, 2022.

\bibitem{weng2019deepchain}
J.~Weng, J.~Weng, J.~Zhang, M.~Li, Y.~Zhang, and W.~Luo, ``Deepchain: Auditable
  and privacy-preserving deep learning with blockchain-based incentive,''
  \emph{IEEE Transactions on Dependable and Secure Computing}, 2019.

\bibitem{rathore2019blockdeepnet}
S.~Rathore, Y.~Pan, and J.~H. Park, ``Blockdeepnet: a blockchain-based secure
  deep learning for iot network,'' \emph{Sustainability}, vol.~11, no.~14, p.
  3974, 2019.

\bibitem{qu2022fl}
Y.~Qu, C.~Xu, L.~Gao, Y.~Xiang, and S.~Yu, ``Fl-sec: Privacy-preserving
  decentralized federated learning using signsgd for the internet of
  artificially intelligent things,'' \emph{IEEE Internet of Things Magazine},
  vol.~5, no.~1, pp. 85--90, 2022.

\bibitem{xu2021bafl}
C.~Xu, Y.~Qu, P.~W. Eklund, Y.~Xiang, and L.~Gao, ``Bafl: an efficient
  blockchain-based asynchronous federated learning framework,'' in \emph{2021
  IEEE Symposium on Computers and Communications (ISCC)}.\hskip 1em plus 0.5em
  minus 0.4em\relax IEEE, 2021, pp. 1--6.

\bibitem{hu2020personalized}
R.~Hu, Y.~Guo, H.~Li, Q.~Pei, and Y.~Gong, ``Personalized federated learning
  with differential privacy,'' \emph{IEEE Internet of Things Journal}, vol.~7,
  no.~10, pp. 9530--9539, 2020.

\bibitem{mcmahan2021advances}
H.~B. McMahan \emph{et~al.}, ``Advances and open problems in federated
  learning,'' \emph{Foundations and Trends{\textregistered} in Machine
  Learning}, vol.~14, no.~1, 2021.

\bibitem{khodak2019adaptive}
M.~Khodak, M.-F.~F. Balcan, and A.~S. Talwalkar, ``Adaptive gradient-based
  meta-learning methods,'' \emph{Advances in Neural Information Processing
  Systems}, vol.~32, 2019.

\bibitem{smith2017federated}
V.~Smith, C.-K. Chiang, M.~Sanjabi, and A.~S. Talwalkar, ``Federated multi-task
  learning,'' \emph{Advances in neural information processing systems},
  vol.~30, 2017.

\bibitem{shlezinger2020communication}
N.~Shlezinger, S.~Rini, and Y.~C. Eldar, ``The communication-aware clustered
  federated learning problem,'' in \emph{2020 IEEE International Symposium on
  Information Theory (ISIT)}.\hskip 1em plus 0.5em minus 0.4em\relax IEEE,
  2020, pp. 2610--2615.

\bibitem{wang2019federated}
K.~Wang, R.~Mathews, C.~Kiddon, H.~Eichner, F.~Beaufays, and D.~Ramage,
  ``Federated evaluation of on-device personalization,'' \emph{arXiv preprint
  arXiv:1910.10252}, 2019.

\bibitem{kim2019blockchained}
H.~Kim, J.~Park, M.~Bennis, and S.-L. Kim, ``Blockchained on-device federated
  learning,'' \emph{IEEE Communications Letters}, vol.~24, no.~6, pp.
  1279--1283, 2019.

\bibitem{kang2019incentive}
J.~Kang, Z.~Xiong, D.~Niyato, S.~Xie, and J.~Zhang, ``Incentive mechanism for
  reliable federated learning: A joint optimization approach to combining
  reputation and contract theory,'' \emph{IEEE Internet of Things Journal},
  vol.~6, no.~6, pp. 10\,700--10\,714, 2019.

\bibitem{li2021blockchain}
J.~Li, Y.~Shao, K.~Wei, M.~Ding, C.~Ma, L.~Shi, Z.~Han, and H.~V. Poor,
  ``Blockchain assisted decentralized federated learning (blade-fl):
  Performance analysis and resource allocation,'' \emph{IEEE Transactions on
  Parallel and Distributed Systems}, vol.~33, no.~10, pp. 2401--2415, 2021.

\bibitem{pokhrel2020federated}
S.~R. Pokhrel and J.~Choi, ``Federated learning with blockchain for autonomous
  vehicles: Analysis and design challenges,'' \emph{IEEE Transactions on
  Communications}, vol.~68, no.~8, pp. 4734--4746, 2020.

\bibitem{lu2020communication}
Y.~Lu, X.~Huang, K.~Zhang, S.~Maharjan, and Y.~Zhang, ``Communication-efficient
  federated learning and permissioned blockchain for digital twin edge
  networks,'' \emph{IEEE Internet of Things Journal}, 2020.

\bibitem{ramanan2020baffle}
P.~Ramanan and K.~Nakayama, ``Baffle: Blockchain based aggregator free
  federated learning,'' in \emph{2020 IEEE International Conference on
  Blockchain (Blockchain)}.\hskip 1em plus 0.5em minus 0.4em\relax IEEE, 2020,
  pp. 72--81.

\bibitem{desai2021blockfla}
H.~B. Desai, M.~S. Ozdayi, and M.~Kantarcioglu, ``Blockfla: Accountable
  federated learning via hybrid blockchain architecture,'' in \emph{Proceedings
  of the Eleventh ACM Conference on Data and Application Security and Privacy},
  2021, pp. 101--112.

\bibitem{mohanta2018overview}
B.~K. Mohanta, S.~S. Panda, and D.~Jena, ``An overview of smart contract and
  use cases in blockchain technology,'' in \emph{2018 9th International
  Conference on Computing, Communication and Networking Technologies
  (ICCCNT)}.\hskip 1em plus 0.5em minus 0.4em\relax IEEE, 2018, pp. 1--4.

\bibitem{cui2020creat}
L.~Cui, X.~Su, Z.~Ming, Z.~Chen, S.~Yang, Y.~Zhou, and W.~Xiao, ``Creat:
  Blockchain-assisted compression algorithm of federated learning for content
  caching in edge computing,'' \emph{IEEE Internet of Things Journal}, 2020.

\bibitem{rahman2020secure}
M.~A. Rahman, M.~S. Hossain, M.~S. Islam, N.~A. Alrajeh, and G.~Muhammad,
  ``Secure and provenance enhanced internet of health things framework: A
  blockchain managed federated learning approach,'' \emph{Ieee Access}, vol.~8,
  pp. 205\,071--205\,087, 2020.

\bibitem{nguyen2021federated}
D.~C. Nguyen, M.~Ding, Q.-V. Pham, P.~N. Pathirana, L.~B. Le, A.~Seneviratne,
  J.~Li, D.~Niyato, and H.~V. Poor, ``Federated learning meets blockchain in
  edge computing: Opportunities and challenges,'' \emph{IEEE Internet of Things
  Journal}, 2021.

\bibitem{ongaro2014search}
D.~Ongaro and J.~Ousterhout, ``In search of an understandable consensus
  algorithm,'' in \emph{2014 $\{$USENIX$\}$ Annual Technical Conference
  ($\{$USENIX$\}$$\{$ATC$\}$ 14)}, 2014, pp. 305--319.

\bibitem{bolosky2011paxos}
W.~J. Bolosky, D.~Bradshaw, R.~B. Haagens, N.~P. Kusters, and P.~Li, ``Paxos
  replicated state machines as the basis of a high-performance data store,'' in
  \emph{Proc. NSDI’11, USENIX Conference on Networked Systems Design and
  Implementation}, 2011, pp. 141--154.

\bibitem{idoje2021survey}
G.~Idoje, T.~Dagiuklas, and M.~Iqbal, ``Survey for smart farming technologies:
  Challenges and issues,'' \emph{Computers \& Electrical Engineering}, vol.~92,
  p. 107104, 2021.

\bibitem{korber2020tracking}
B.~Korber, W.~M. Fischer, S.~Gnanakaran, H.~Yoon, J.~Theiler, W.~Abfalterer,
  N.~Hengartner, E.~E. Giorgi, T.~Bhattacharya, B.~Foley \emph{et~al.},
  ``Tracking changes in sars-cov-2 spike: evidence that d614g increases
  infectivity of the covid-19 virus,'' \emph{Cell}, vol. 182, no.~4, pp.
  812--827, 2020.

\bibitem{li2020privacy}
Y.~Li, Y.~Zhou, A.~Jolfaei, D.~Yu, G.~Xu, and X.~Zheng, ``Privacy-preserving
  federated learning framework based on chained secure multi-party computing,''
  \emph{IEEE Internet of Things Journal}, 2020.

\bibitem{deng2009imagenet}
J.~Deng, W.~Dong, R.~Socher, L.-J. Li, K.~Li, and L.~Fei-Fei, ``Imagenet: A
  large-scale hierarchical image database,'' in \emph{2009 IEEE conference on
  computer vision and pattern recognition}.\hskip 1em plus 0.5em minus
  0.4em\relax Ieee, 2009, pp. 248--255.

\bibitem{reyes2012human}
D.~Anguita, A.~Ghio, L.~Oneto, X.~Parra, J.~L. Reyes-Ortiz \emph{et~al.}, ``A
  public domain dataset for human activity recognition using smartphones.'' in
  \emph{Esann}, vol.~3, 2013, p.~3.

\bibitem{sztyler2016body}
T.~Sztyler and H.~Stuckenschmidt, ``On-body localization of wearable devices:
  An investigation of position-aware activity recognition,'' in \emph{2016 IEEE
  International Conference on Pervasive Computing and Communications
  (PerCom)}.\hskip 1em plus 0.5em minus 0.4em\relax IEEE, 2016, pp. 1--9.

\bibitem{rajagopalan2022deep}
C.~Rajagopalan, D.~Rawlinson, E.~Goldberg, and G.~Kowadlo, ``Deep learning in a
  bilateral brain with hemispheric specialization,'' \emph{arXiv preprint
  arXiv:2209.06862}, 2022.

\bibitem{zhang2020democratically}
T.~Zhang, Z.~Shen, J.~Jin, and X.~Zheng, ``A democratically collaborative
  learning scheme for fog-enabled pervasive environments,'' in \emph{2020 IEEE
  International Conference on Pervasive Computing and Communications Workshops
  (PerCom Workshops)}.\hskip 1em plus 0.5em minus 0.4em\relax IEEE, 2020, pp.
  1--4.

\end{thebibliography}
\bibliographystyle{IEEEtran}

\begin{IEEEbiography}[{\includegraphics[width=1in,height=1.25in,clip,keepaspectratio]{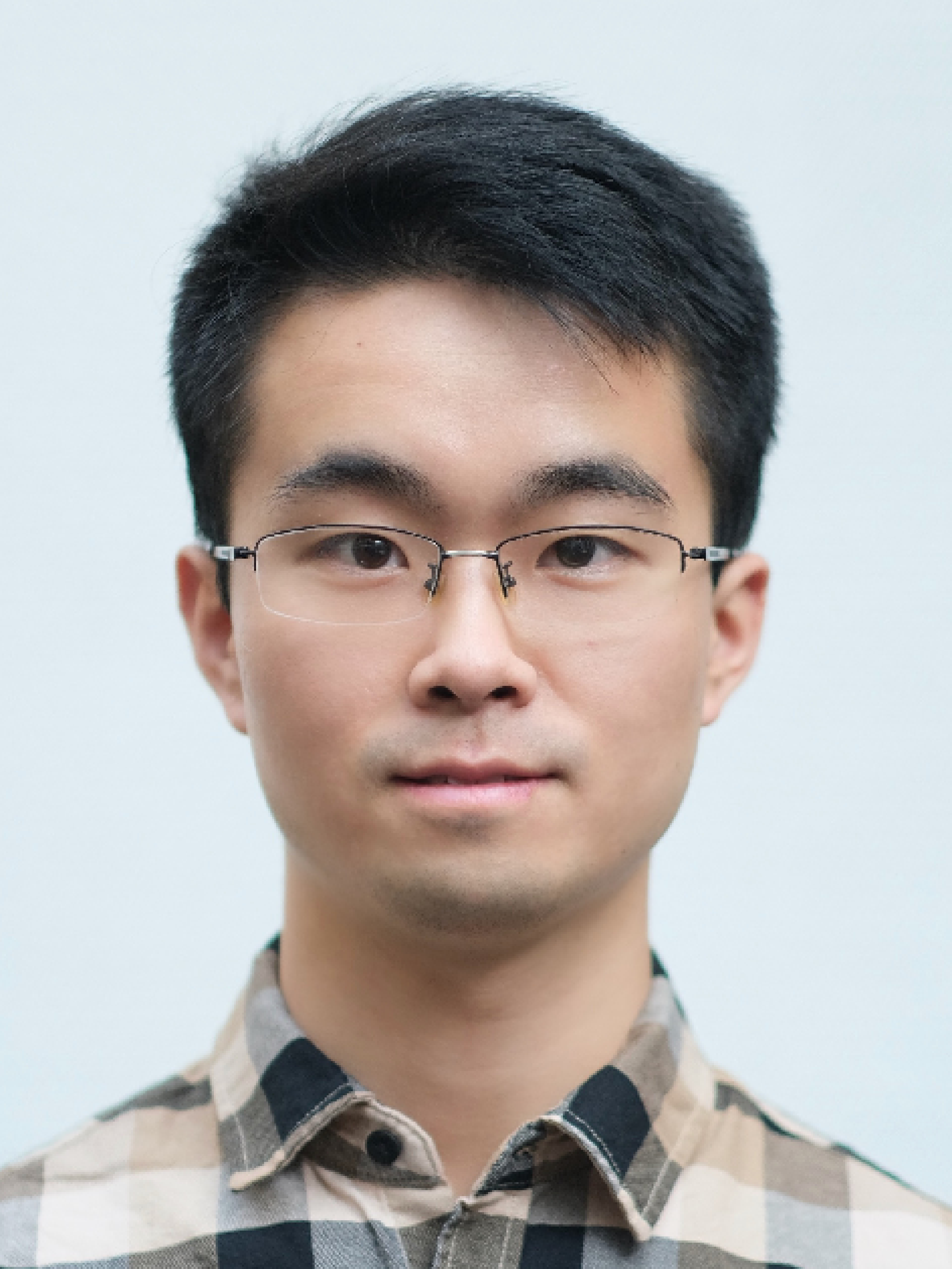}}]{Chenhao~Xu} is an Associate Research Fellow at the School of Information Technology, Deakin University. He received the BS degree in Software Engineering from Beijing Institute of Technology, China, in 2018 and the PhD degree in Information Technology from Deakin University, Australia, in 2023. He served as the Web Chair and Technical Program Committee Member of EAI TRIDENTCOM 2022. He also serves as a Review Editor of Frontiers in Big Data. His research interests include blockchain, federated learning, and edge computing.
\end{IEEEbiography}

\begin{IEEEbiography}[{\includegraphics[width=1in,height=1.25in,clip,keepaspectratio]{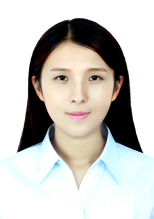}}]{Jiaqi~Ge} received the B.S. and M.S. degree from the College of Computer Science and Technology, Jilin University, China, where she is currently pursuing the Ph.D. degree. Her main research interest is distributed computing, and now she's currently working on federated learning and blockchain.
\end{IEEEbiography}

\begin{IEEEbiography}[{\includegraphics[width=1in,height=1.25in,clip,keepaspectratio]{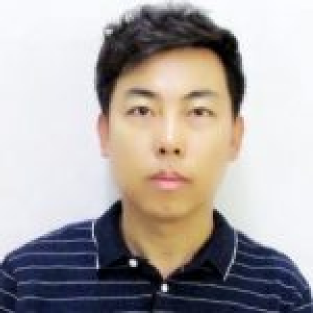}}]{Yong~Li} received the M.S. degree in the Architecture of Computer System from the Jilin University in 2004. He is currently pursuing the PhD degree with Jilin University, China. He is an associate professor of the Network Engineering in Changchun University of Technology, China. His research interests include Federated Learning, Egde-computing, Information Security and Privacy-preserving.
\end{IEEEbiography}

\begin{IEEEbiography}[{\includegraphics[width=1in,height=1.25in,clip,keepaspectratio]{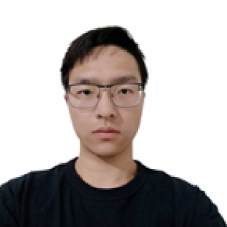}}]{Yao~Deng} is a PhD candidate at the Macquarie University, Australia. He received the Bachelor degree of Information Technology from the Deakin University, Australia in March, 2018, the Bachelor degree of Software Engineering from the SouthWest University, China in July, 2018, and the Master of Research degree from the Macquarie University. He is currently a IEEE student member. His current research interests include adversarial attacks and defenses, testing and anomaly detection of deep learning based autonomous driving systems.
\end{IEEEbiography}

\begin{IEEEbiography}[{\includegraphics[width=1in,height=1.25in,clip,keepaspectratio]{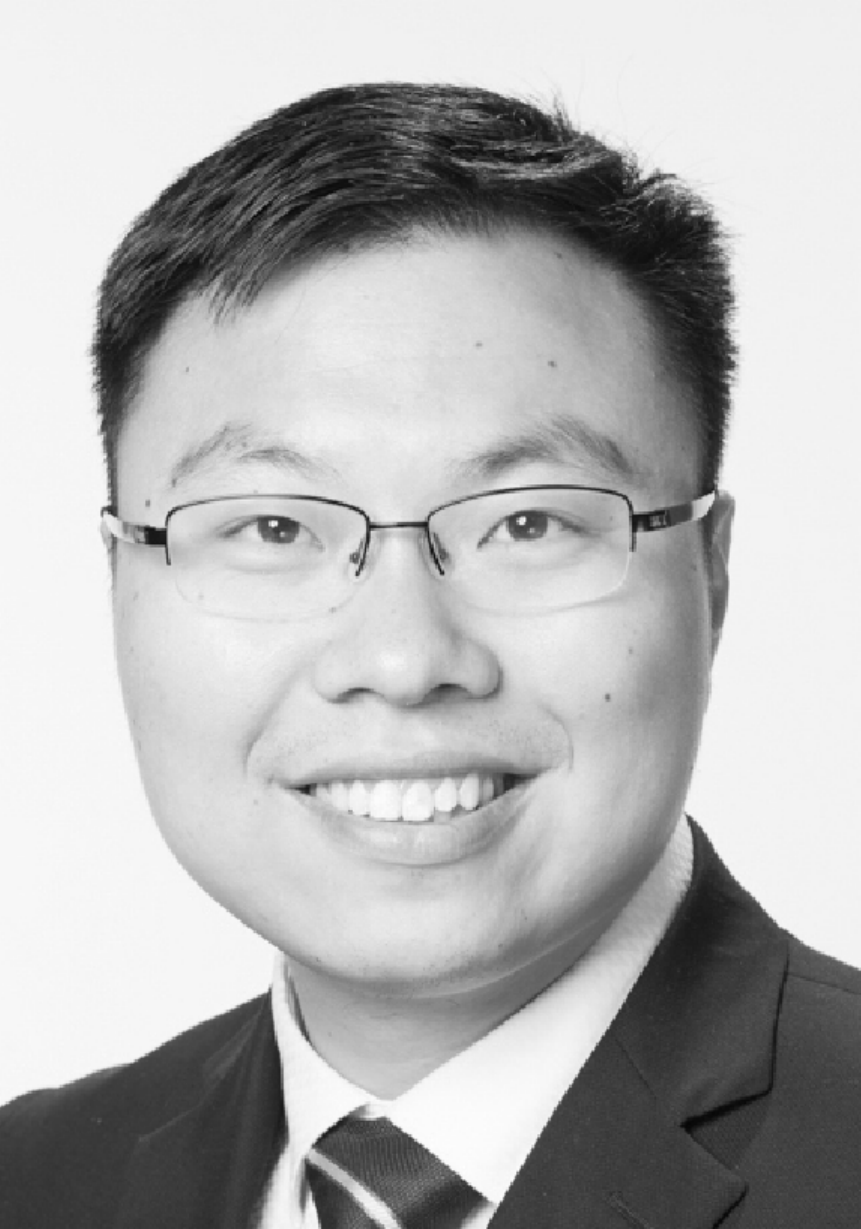}}]{Longxiang~Gao} (SM17) received his PhD in Computer Science from Deakin University, Australia. He is currently a Professor at Qilu University of Technology (Shandong Academy of Sciences) and Shandong Computer Science Center (National Supercomputer Center in Jinan). He was a Senior Lecturer at School of Information Technology, Deakin University and a post-doctoral research fellow at IBM Research \& Development, Australia. His research interests include Fog/Edge computing, Blockchain, data analysis and privacy protection.

Dr. Gao has over 90 publications, including patent, monograph, book chapter, journal and conference papers. Some of his publications have been published in the top venue, such as IEEE TMC, IEEE TPDS, IEEE IoTJ, IEEE TDSC, IEEE TVT, IEEE TCSS, IEEE TII and IEEE TNSE. He has being Chief Investigator (CI) for more than 20 research projects (the total awarded amount is over \$5 million), from pure research project to contracted industry research. He is a Senior Member of IEEE.
\end{IEEEbiography}

\begin{IEEEbiography}[{\includegraphics[width=1in,height=1.25in,clip,keepaspectratio]{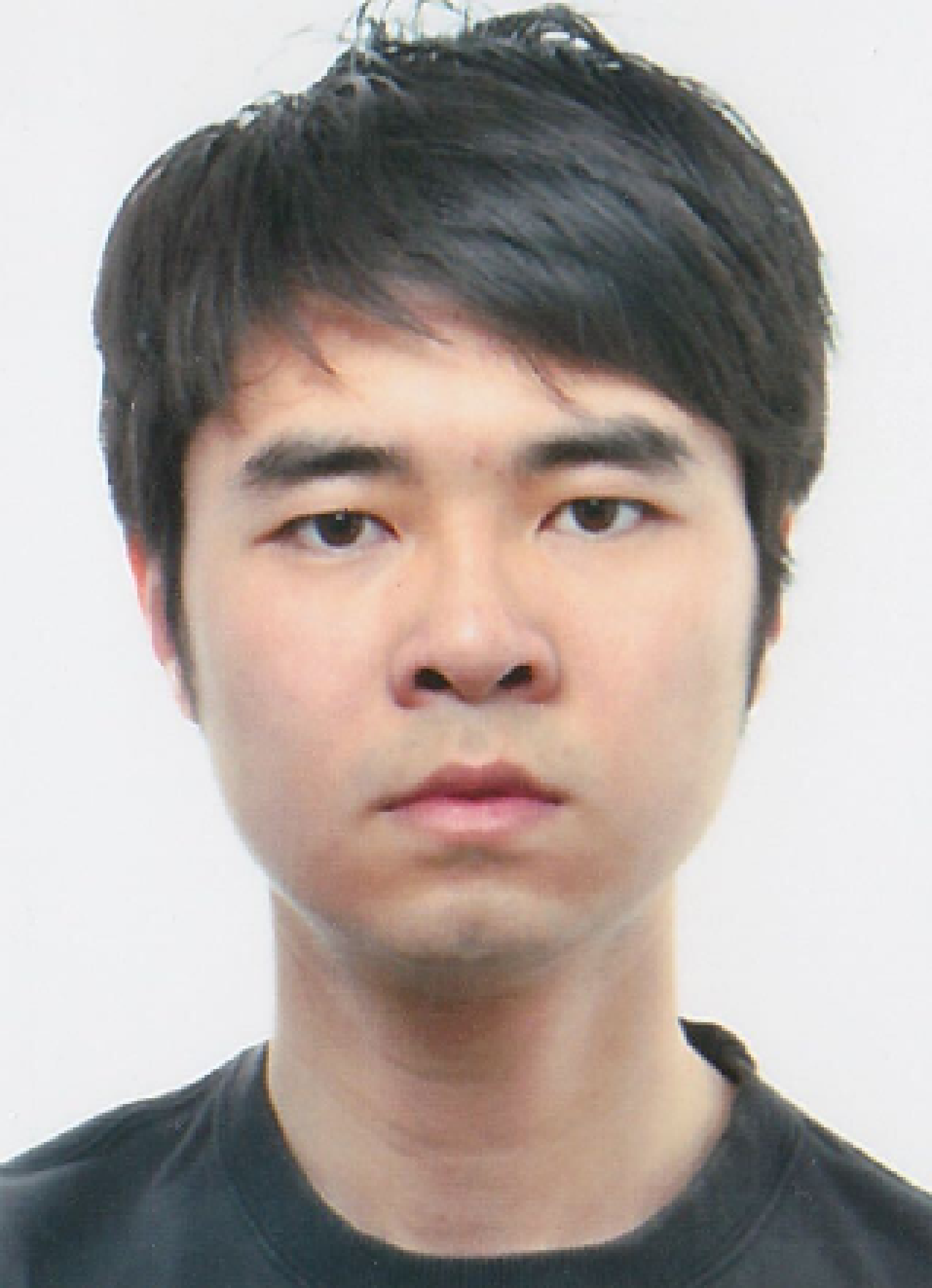}}]{Mengshi~Zhang} received the BS degree in electronic engineering from Tsinghua University, in July 2014 and the PhD degree from the Department of Electrical and Computer Engineering, University of Texas at Austin, in August 2019. His research interests include fault localization, program repair, and machine-learning-oriented software engineering.
\end{IEEEbiography}

\begin{IEEEbiography}[{\includegraphics[width=1in,height=1.25in,clip,keepaspectratio]{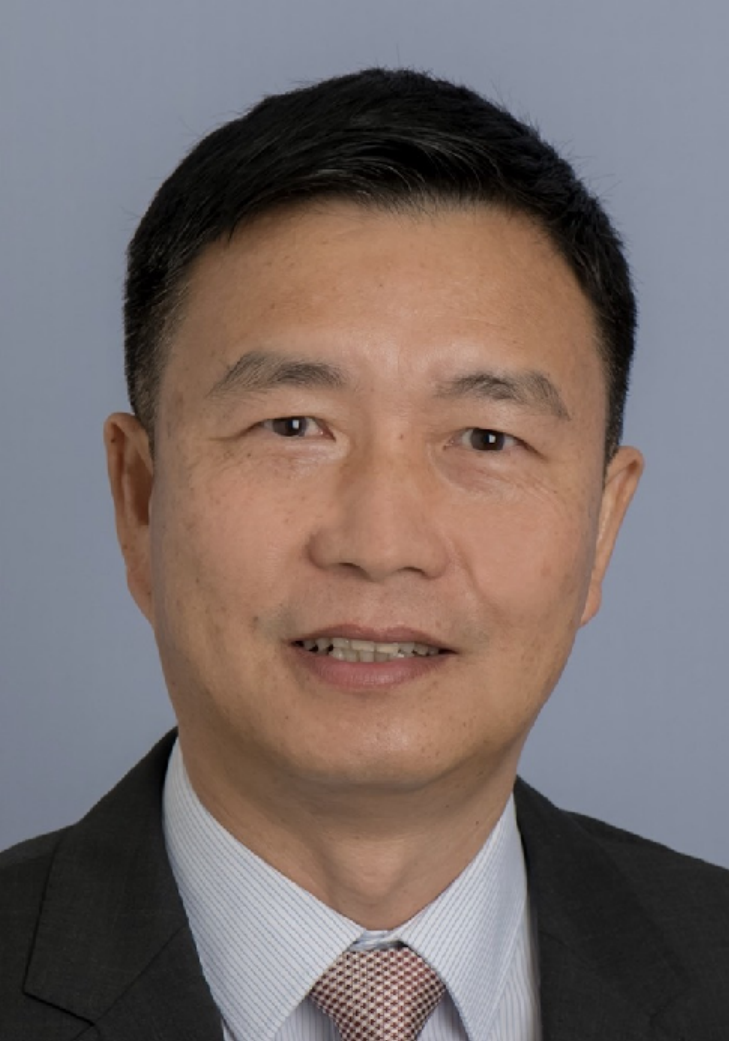}}]{Yong~Xiang} received his B.E. and M.E. degrees from the University of Electronic Science and Technology of China, China, and PhD degree from The University of Melbourne, Australia. He is a Professor at the School of Information Technology, Deakin University, Australia. He is also the Associate Head of School (Research) and the Director of the Artificial Intelligence and Data Analytics Research Cluster. He has obtained a number of research grants (including several ARC Discovery and Linkage grants from the Australian Research Council) and published numerous research papers in high-quality international journals and conferences. He is the coinventor of two U.S. patents and some of his research results have been commercialised. Dr Xiang is the Editor/Guest Editor of several international journals. He has been invited to give keynote speeches and chair committees in a number of international conferences, review papers for many international journals and conferences, serve on conference program committees, and chair technical sessions in conferences. Dr. Xiang is a senior member of the IEEE.
\end{IEEEbiography}

\begin{IEEEbiography}[{\includegraphics[width=1in,height=1.25in,clip,keepaspectratio]{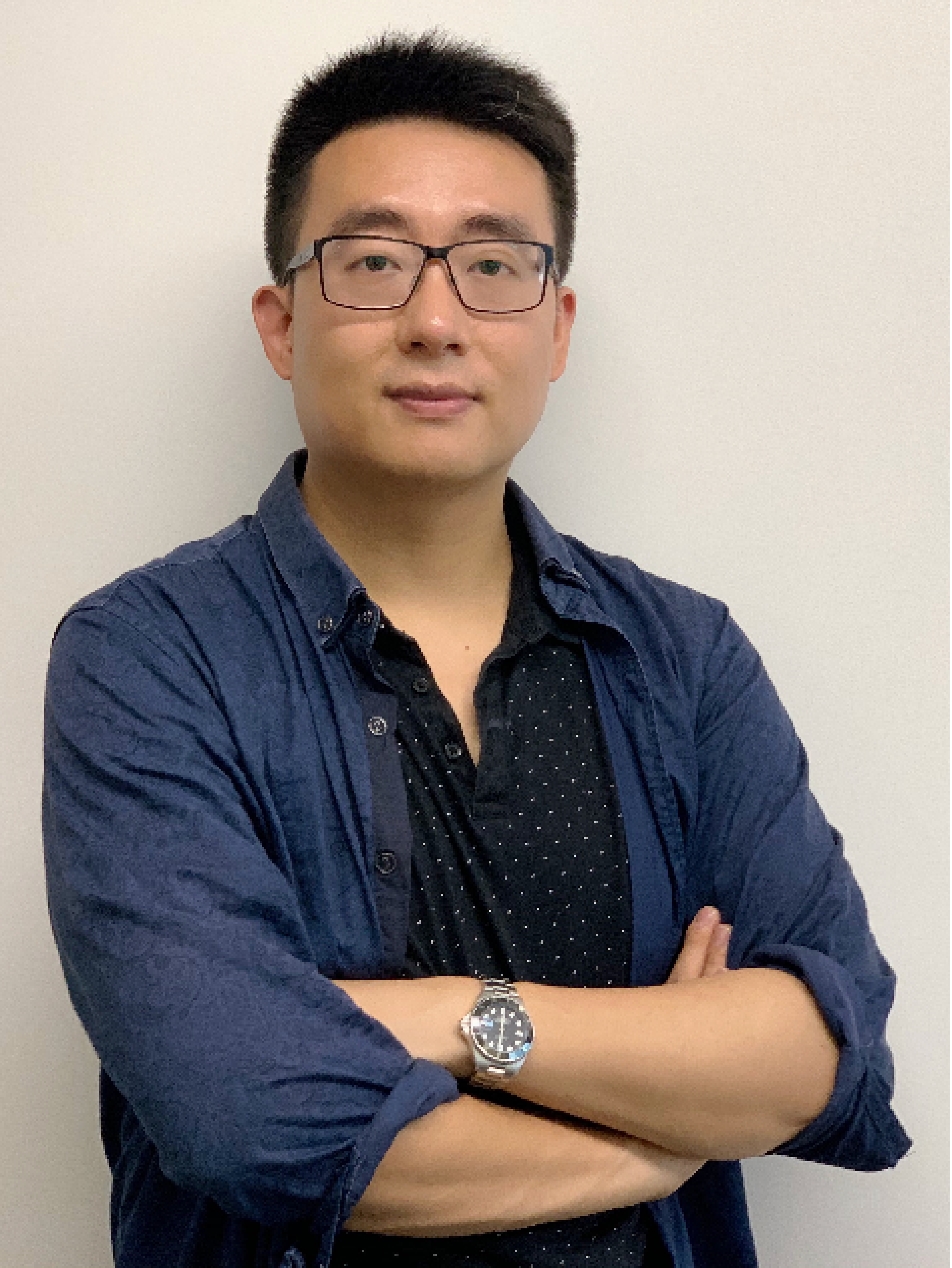}}]{Xi~Zheng} received the Ph.D. in Software Engineering from UT Austin in 2015. From 2005 to 2012, he was the Chief Solution Architect for Menulog Australia. He is currently the Director of Intelligent Systems Research Group (ITSEG.ORG), Senior Lecturer (aka Associate Professor US) and Deputy Program Leader in Software Engineering, Macquarie University, Sydney, Australia.  His research interests include CPS Verification, Machine Learning Security, Human Vehicle Interaction, Edge Intelligence and Intelligent Software Engineering. He has a number of highly cited papers and serves as PC members for PerCom (CORE A*) and TrustCom (CORE A).
\end{IEEEbiography}

\end{document}